\documentclass[wcp,openany]{jmlr}

\usepackage{longtable}

\usepackage{booktabs}

\usepackage{multirow}

\makeatletter
\let\Ginclude@graphics\@org@Ginclude@graphics 
\makeatother

\title{Hierarchical Semantic Segmentation using Psychometric Learning}

  \author{\Name{Lu Yin} \Email{l.yin@tue.nl}\\
   \Name{Vlado Menkovski} \Email{V.Menkovski@tue.nl}\\
   \Name{Shiwei Liu} \Email{s.liu3@tue.nl}\\
   \Name{Mykola Pechenizkiy} \Email{m.pechenizkiy@tue.nl}\\
  \addr Eindhoven University of Technology, Eindhoven 5600 MB, Netherlands}

\begin{document}

\maketitle

\begin{abstract}
Assigning meaning to parts of image data is the goal of semantic image segmentation. Machine learning methods, specifically supervised learning is commonly used in a variety of tasks formulated as semantic segmentation. One of the major challenges in the supervised learning approaches is expressing and collecting the rich knowledge that experts have with respect to the meaning present in the image data. Towards this, typically a fixed set of labels is specified and experts are tasked with annotating the pixels, patches or segments in the images with the given labels. In general, however, the set of classes does not fully capture the rich semantic information present in the images. For example, in medical imaging such as histology images, the different parts of cells could be grouped and sub-grouped based on the expertise of the pathologist. 

To achieve such a precise semantic representation of the concepts in the image, we need access to the full depth of knowledge of the annotator. In this work, we develop a novel approach to collect segmentation annotations from experts based on psychometric testing. Our method consists of the psychometric testing procedure, active query selection,  query enhancement, and a deep metric learning model to achieve a patch-level image embedding that allows for semantic segmentation of images. We show the merits of our method with evaluation on the synthetically generated image, aerial image and histology image.

\end{abstract}
\begin{keywords}
Psychometric test, Hierarchical Semantic Segmentation, Deep metric learning, Active learning.
\end{keywords}

\begin{figure}[htbp]
    \centering 

    \subfigure[
Hierarchical image segmentation by contour detection
]{
        \includegraphics[width=0.76\textwidth]{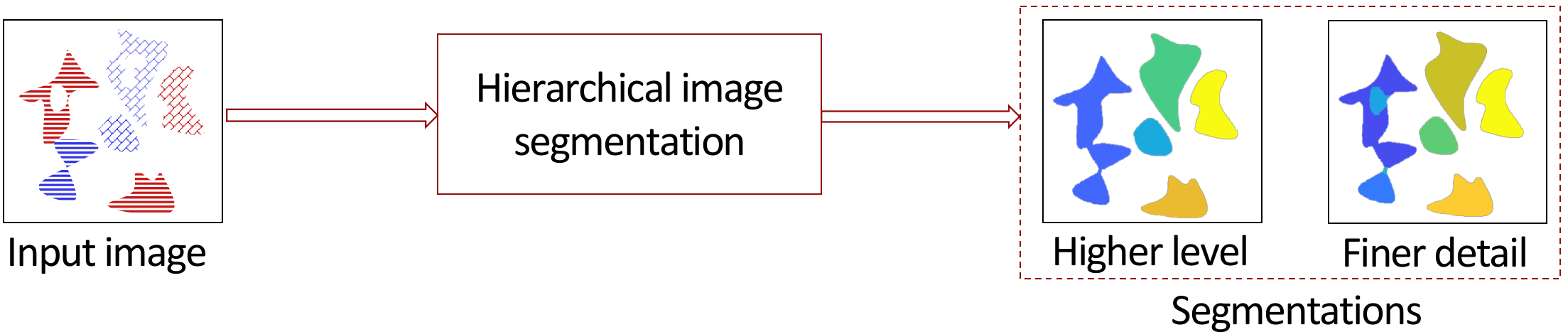}
        \label{Sys_compare_1}
    }
    
    \subfigure[Supervised segmentation with 
pre-defined classes
]{
        \includegraphics[width=0.6\textwidth]{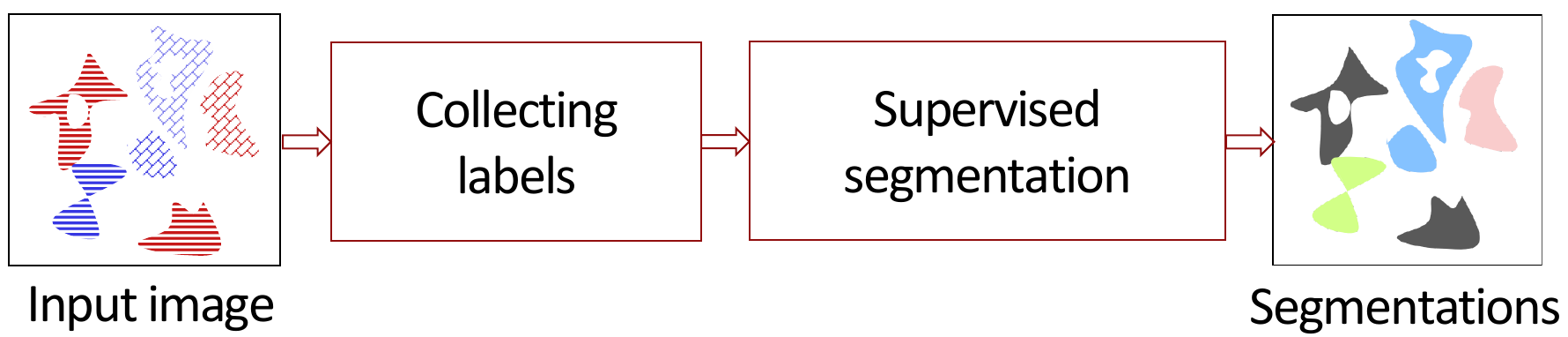}
        \label{Sys_compare_2}
    }

    \subfigure[Proposed framework]{
        \includegraphics[width=1\textwidth]{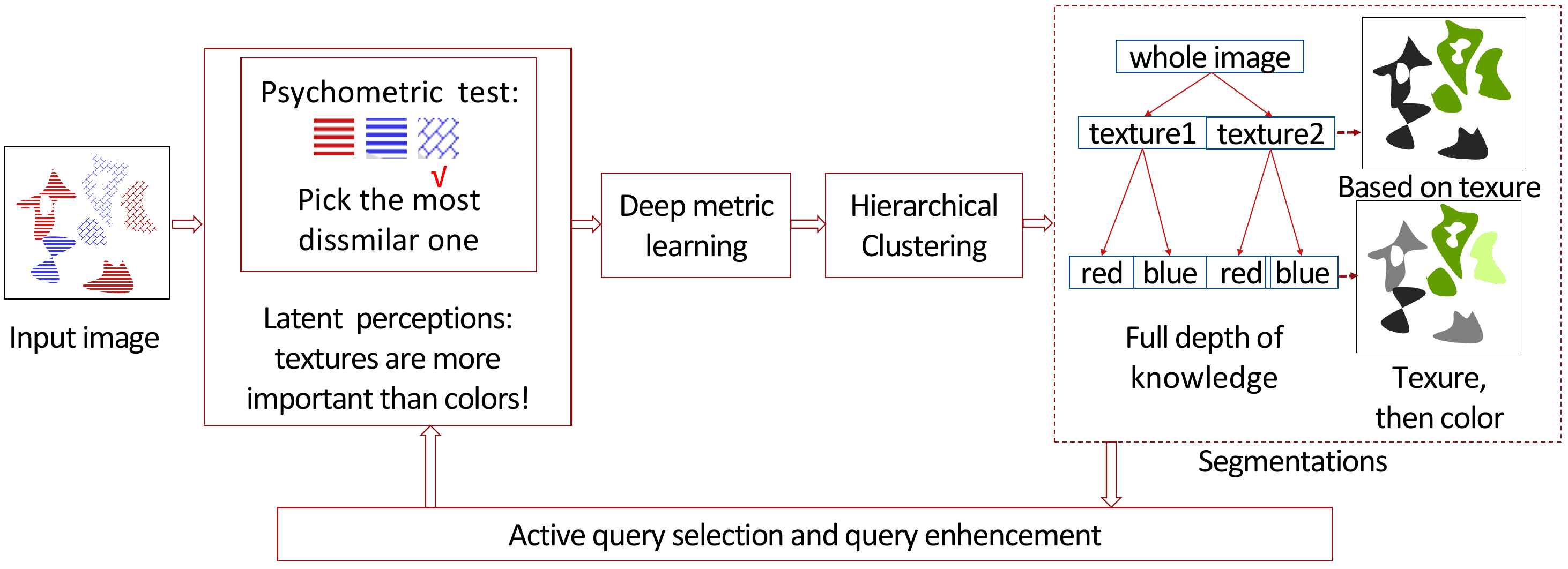}
        \label{Sys_whole}
    }

\caption{A comparison of three different segmentation methods using an example image containing six regions with two different textures and two different colors. (a):  \textit{Hierarchical image segmentation by contour detection}. Segmentation is implemented only based on information present in the image. Domain knowledge can not be utilized to form complex assignments.
(b): \textit{Supervised segmentation}. A fixed set of labels is assigned to each pixel. Coarse-level domain knowledge is expressed, but high fidelity   hierarchical relationships present in the image are discarded.
(c): \textit{Our proposed framework}. Psychometric tests are applied to collect full-depth knowledge from the domain experts. In the test, three patches are sampled from the image for comparison. We assume the domain experts tend to choose the third option (blue cross slash) as the most dissimilar one since they perceive textures as more important than colors. After collecting a number of test responses, the image could be hierarchically segmented based on the texture on a higher level and then divided by colors, which reflects the captured experts' perception structure. Besides, concept relationships are mapped to color similarities for visually inspecting. For instance, in the second level segmentation, two different but similar colors, dark green and light green, represent two also similar but not the same concepts, red cross slash and blue cross slash. Note that the square patches in the test are just for illustration and are replaced by super-pixels in the following experiments. }


    \label{Compare}
\end{figure}

\section{Introduction}

One of the major challenges in the task of image semantic segmentation is collecting high-quality supervision from the domain expert. Typically this task is reduced to assign a label to an image region from a fixed set of labels~\citep{vernaza2017learning,wang2020self,khoreva2017simple}. While in some cases, such reduction may have sufficient semantic fidelity in general, this is still a major limitation. One can imagine that in the majority of applications, the concepts present in the images have a much more complex relationship between each other and possibly form some kind of hierarchies. For example, when dealing with histology image segmentation, rather than forcing the expert to project their domain knowledge to flat labels (e.g., in the case of biopsy segmentation, one set of classes may include: normal, benign, in-situ and invasive \citep{aresta2019bach}), we would let them express their full-depth knowledge such that the experts can express the subtle differences in the contents of any two patches.
Some scholars tried to use hierarchical segmentation method~\citep{arbelaez2010contour} to capture the pattern relationships in an image. As no semantic supervision is provided from the domain expert, the built hierarchical structure is still not able to match  the full-depth expert knowledge.

To address these challenges, we developed a psychometric-test-based approach to elicit the expert's high fidelity knowledge, and map it to semantic hierarchical segmentation. Firstly, we create super-pixel patches by the SLIC~\citep{achanta2012slic} approach and use psychometric test to measure the patches' perceived similarity to each other. Next, we apply a deep metric learning algorithm to project the perceived relative distances to an embedded space where we can structure the patches in a hierarchical fashion, thereby proceed to image hierarchical segmentation. The built hierarchy reflects the latent perception structure, which we aim to capture from the annotator. To visualize the structure, we assign the segmented regions different color overlays from a given palette. The distance of the colors in the palette corresponds to the distances between patches in the embedded space. This allows us to visually inspect not only the captured relationships between the concepts of the image in the embedded space but also concretely the image.

Our proposed approach is illustrated in a toy example and compared with two baselines in Fig~\ref{Compare}. It is shown that our method could capture the full-depth knowledge from experts and form a hierarchical semantic segmentation accordingly.

The rest of our paper is organized as follows: in section 2 we review the most related researches to our work; section 3 explains the details of our approach; in section 4 we evaluate our experiments in three different scenarios, carry an ablation study, and compare our approach with two baselines quantitatively; at last, we conclude our work and contributions in section 5.


\section{The Related Work}

There are three research lines closely related to our work:  psychometric test, semantic image segmentation and deep metric learning.

\textbf{Psychometric Test}
Psychometric test studies the perceptual processes of psychical stimuli~\citep{gescheider2013psychophysics}. It has a long researched history and wide applications in different areas~\citep{son2006x,feng2014methodology}.  Some scholars use discriminative-based psychometric tests to measure the slightest difference between two stimuli that a subject can detect. It could provide more accrue results with fewer noises than directly assigning quantified values~\citep{jogan2014new}. Especially, the two-alternative-force choice (2AFC) model was developed in~\cite{fechner1860elemente}, in which subjects are asked to compare the differences between two perceived stimuli and forced to make a correct choice.  For example, two cups of water are given, the participant is asked to compare which one is sweeter.  This experiment has been adopted to measure the subjects' perception of more complex multimedia content such as images or videos~\citep{son2006x,feng2014methodology}. As there are only two options in the 2AFC test, limited perception is captured by a test and some of the questions might be ambiguous to subjects. Some scholars extend the compared objects to $m$, forming the M-AFC family~\cite{decarlo2012signal}. In this work, we use a variation of 2AFC, the three-alternative-force choice(3AFC) test, in which three samples are compared to capture the annotator's perception of images.

\textbf{Semantic image segmentation} Based on the type of annotation, image segmentation could be divided into the supervised and semi-supervised way.  The former one required pixel-level class labels, which is laborious but achieving better performance.  FCN is a milestone for supervised segmentation and inspired many subsequent researches~\citep{long2015fully}.  It provided an end-to-end trainable model and available for the arbitrary size of inputs. Parsenet~\citep{liu2015parsenet} extended FCN by considering the global context information. Atrous convolution was integrated into the Deeplab segmentation family~\citep{chen2018encoder} to enlarge the receptive field without increasing the computational cost. Attention-mechanisms-based segmentation was explored to assign different weights to objects in~\citet{fu2019dual}. Weakly supervised semantic segmentation (WSSS), on the other hand, uses weak labels instead of pixel-level annotations to guide model training, try to achieve a balance between performance and annotation costs. Different forms of supervision have been explored: image-level classification labels~\citep{wang2020self}, bounding boxes~\citep{khoreva2017simple} or scribbles~\citep{vernaza2017learning}. 

However, these researches still suffer from the limitation of reducing the expert's high-quality supervision to a fixed set of labels. In this paper, we apply the psychometric learning based way to collect full-depth knowledge from experts.

\textbf{Deep metric learning and informative pair selection}
Metric learning aims to learn a measure of distance between different data points and has experienced great progress due to the development of deep learning~\citep{kaya2019deep}. Many approaches are developed by the comparison-based model. Contrastive loss was applied to force a margin distance between positive and negative pairs by making pair-wise comparison~\citep{hadsell2006dimensionality}.  triplet-net~\citep{schroff2015facenet}, quadruplet-net~\citep{chen2017beyond} and N-pair-net ~\citep{sohn2016improved} extend the compared sample numbers  in a loss-function to three, four and $N$.  Though yielding promising results, it is impractical to exhaust all the possible pair combinations for model training. Different approaches were proposed to select nontrivial samples.    Semi-hard was applied in \citet{schroff2015facenet} to select triplets that conflict with the loss function.  \citet{suh2019stochastic} use class-based strategy to mining sample in a coarse-to-fine fashion.  The most related work is from \citet{yin2020knowledge}. The scholars developed a dual-triplet loss function in which only negative image needs to be specified, and they applied a Bayesian-based scheme to actively select triplet queries when an actual human participant is involved.  We followed that work's spirit and discussed the impacts of different margin settings. Besides, we extend the query selection method to the segmentation scenario and propose a query enhancement strategy to improve the query efficiency further.

\section{Approach}\label{Approach}
As shown in Fig~\ref{Sys_whole}, our psychometric learning based semantic segmentation framework consists of four parts, psychometric test, deep metric learning, hierarchical clustering, and active query selection with query enhancement.

First, we use psychometric tests to elicit the annotator's perception. Detailly, we use discriminative-based psychometric testing to perceive relative differences among super-pixel patches created by SLIC~\citep{achanta2012slic}. Our rich semantic knowledge could be mapped to a model by performing simple comparison tasks without assigning specific labels to image patches.

Then, a deep metric learning method is applied to project the perceived relative distances to embedding distances. That is, if an annotator believes two image patches are more similar than others, their relative distance should also be closer in the embedding space. Similar to ~\citet{yin2020knowledge}. We use the dual-triplet loss to align with our psychometric tests.

Next, in the embedding space, we cluster and sub-cluster super-pixel patches, organize them in a hierarchical fashion to segment images at different levels. The built hierarchy reflects the captured full-depth knowledge from the annotator. Unlike the hierarchical image segmentation research in~\citet{arbelaez2010contour}, our segmentation is created based on the annotator's responses instead of pure image information. Therefore, we are able to create various hierarchical segmentation based on different annotators' senses.

To visualize the segmentation structure, we assign the segmented regions different colors from a given palette. Since these colors are projected from embedding space that represents the concepts' relationships, the concept similarities could be inspected by colors intuitively.

While the effectiveness of psychometric testing, there could be too many potential queries due to the combination. It is crucial to develop a query selection scheme to select a fraction of proper queries so that the psychometric testing could be carried in an efficient way. Our work uses a Bayesian-based method to select queries with high uncertainty (informative) and high utility (un-ambiguous). Besides, a query augmentation method is applied to simulate more responses by the answered ones.

\subsection{Psychometric Test}
\begin{figure}[htbp]
    \centering 
    \includegraphics[width=0.65\textwidth]{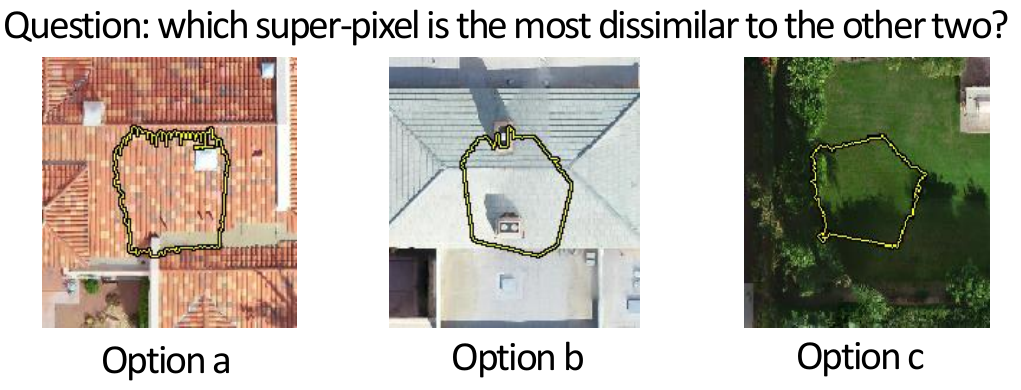} 
    \caption{An example of three-alternative-forced-choice in the aerial image. The subject is asked to compare the super-pixel patches (patches circled by the yellow line) and choose the most dissimilar one against the other two. Surrounding areas are also presented for reference. } 
    \label{3AFC}
\end{figure}

Three-alternative-force choice (3AFC)~\citep{decarlo2012signal} tests are applied to elicit the annotator's perception of an image. To be specific, we create super-pixel patches from an image by SLIC method, sample three from the super-pixels in every test, and ask the annotators to choose the most dissimilar one. By making this simple decision,  perceptions of the patches' relative similarities are elicited. Note that, in the 3AFC test, besides super-pixels themselves, we present their surrounding areas as well to take advantage of more global image information.  An example is shown in Fig.~\ref{3AFC}.

\subsection{Deep Metric Learning}

After psychometric testing, we use deep metric learning to map the captured patch similarities to embedding distances. There are varieties of deep metric learning models available, but a natural choice would be triplet-net~\citep{schroff2015facenet}, as we also have three samples in the 3AFC test. The typical loss function in triplet-net is shown as,

\begin{equation}
L=\sum_{i=1}^N \left[   \|f(x_{a}^i)- f(x_{p}^i)\| - \|f(x_{a}^i)-f(x_{n}^i)\| +m \right]_+ 
\label{o_loss}
\end{equation}
where $x_{a}^i,x_{p}^i$ are anchor and positive images from the same class. $x_{n}^i$ is the negative image from a different class.  $f(\cdot) \in\mathbb{R}^d$ is the learned embedding representation from a neural network. It embeds an image  to a d-dimension euclidean space. $\left[ \cdot \right]_+$   denotes a  hinge function  $max[0,\cdot]$. This loss function enforce a distances margin $m$ between images pairs with same labels to other image pairs.

Note that, in equation~\ref{o_loss}, anchor, negative and positive three images all have to be indicated. However, in a 3AFC test, only the most dissimilar one needs to be specified. To deal with this conflict, \citet{yin2020knowledge}  proposed a dual-triplet loss function which define the chosen image  in 3AFC as negative $x_{n}$ and  rest two images all as  positive images $x_{p1}^i, x_{p2}^i$. This loss introduces a dynamic margin to encourage the image pairs that the annotator perceives similarly to be closer than other pairs in embedding space. In our work, we use a constant margin $m$ instead of a dynamic one and rewrite the dual-triplet  loss as $L_{dual}=$


\begin{small}\label{Triplet_dual}
\begin{equation}
 \!\sum_{i=1}^N \!\left\{ \! \left[  \! \|f(x_{p1}^i) \!-\! f(x_{p2}^i)\| \! -\! \|f(x_{p1}^i) \!- \!f(x_{n}^i)\|\! +\!m \right]_+  \!+\!  \left[  \! \|f(x_{p1}^i) \!-\! f(x_{p2}^i)\| \!- \!\|f(x_{p2}^i) \!- \!f(x_{n}^i)\| \!+ \!m \right]_+ \!\right\}  \label{n_loss}
\end{equation}
\end{small}

The reason for using a constant margin is that the elicited knowledge structure in segmentation is usually not that deep and complex, and a constant margin would be enough to build a discriminative embedding space without extra computation costs. Besides, the extra variables from by dynamic margin could introduce unstable factors. The details of different margin settings' impacts are reported in the Appendix~\ref{appendix}.

\subsection{Hierarchical Clustering}

Once the semantic embedding is obtained, we could hierarchically cluster super-pixel patches in the embedding space to create hierarchical segmentation.

In general, there are two main categories of the hierarchical clustering algorithms, a  bottom-up approach called ``agglomerative" and a top-down approach called ``divisive" ~\citep{rokach2005clustering}.  The agglomerative approach starts with each sample belonging to one single cluster and merges samples together as the hierarchy moves up.
The divisive approach starts with all data as one cluster and splits recursively when the hierarchy goes down. Commonly this approach uses heuristics to choose splits, such as K-means. 

In the agglomerative method, we need to carefully indicate the cutting value when we need a node containing more than two branches in the hierarchy. The cutting value is not intuitive compared with $K$ in K-means because $K$ could represent the number of branches directly.  Therefore,  in our work, we applied the divisive approach with K-means to create hierarchical segmentation. Silhouette score~\citep{kodinariya2013review} is applied to help us decide the cluster number $K$ and when to stop divide.

We also map the concept relationships to color similarities for visually inspecting. In detail, we use MDS~\citep{ghodsi2006dimensionality} reducing the image's embedding dimension to three and scale the embedding value to $0\sim255$ to represent RGB values.  The concepts' relative distances are then mapped to a color space.  For instance, in a histopathological image, the color of in-situ areas should more like invasive areas than others if these two areas' concepts are also more similar in the annotator's perception.

\subsection{Active Query Selection and Query Enhancement}
In psychometric tests, there are $\binom{B}{3}$ potential questions given $B$ patches by combination. Since it is impractical to answer all the queries and not every query contributes equal gradients for training, we developed a scheme to actively select proper queries with uncertainty and utility. Furthermore,  we enhance the answered queries by simulating more responses.

Our selection procedure follows an iterative process. In each iteration, there are three steps.  Our model and the elicited hierarchical knowledge will be updated by the selected queries in each iteration.

First, we generate potential questions by random sampling at each level of the hierarchy created from the last iteration. That allows the model to capture the global distribution as well as the local information in finer detail.

Secondly, inside the potential questions,  we future select queries with high uncertainty.  That is,  queries that annotators are not confident with. Those queries are usually more informative and produce more training gradients~\citep{settles2009active}. We use Bayes' theorem to calculate the distribution of answering possibility and measure the uncertainty by the answers' expected values.

To be specific, for a 3AFC query $q_i$,  we denote the possibilities of picking three options $a,b,c$  as $\theta= (\theta_1,\theta_2,\theta_3)$, and  assume they follows dirichlet's distribution  $Dir(\alpha) $. The distribution density can be written as, 

\begin{equation}
p(\theta|\alpha)=\frac{1}{Beta(\alpha)}\prod_{i=1}^3 \theta_i^{\alpha_{i-1}} (\theta_i\geq0; \sum_{i=1}^3\theta_i=1), Beta(\alpha)=\frac
{ \prod_{i=1}^3 \Gamma (\alpha_i) }{\Gamma( \sum_{i=1}^3 \alpha_i )}
\end{equation}

\begin{figure}[htbp]
    \centering 

    \includegraphics[width=0.45\textwidth]{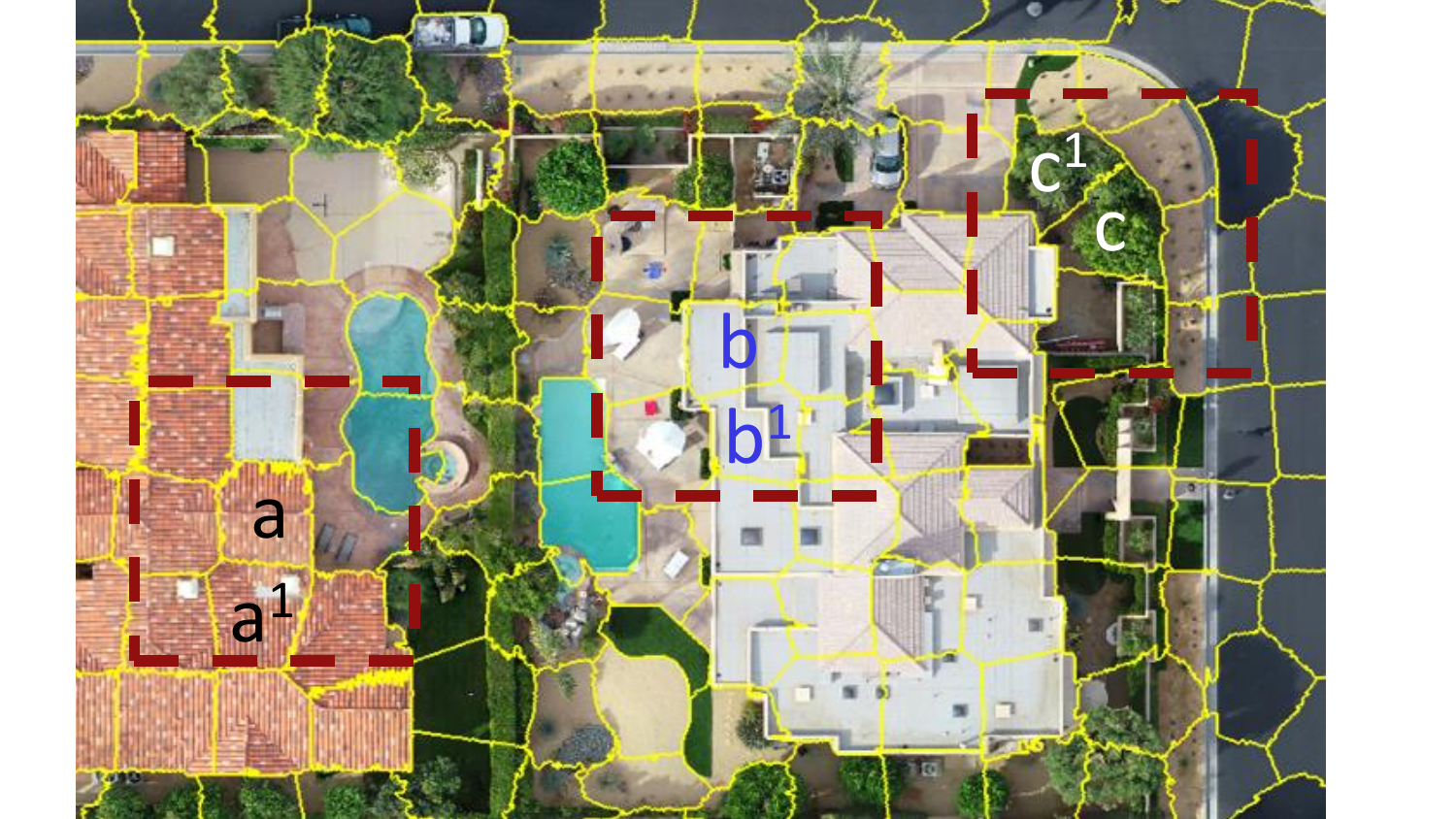} 
    \caption{A query and its most similar query. $a,b,c$ are three super-pixels from a query. In each sample's  neighbour area (bounded by red dotted line), we pick the closest super-pixels $a_1,b_1,c_1$  by their embedding distances to $a,b,c$.  Query [$a_1,b_1,c_1$] then is the similar query to [$a,b,c$].
  } 
    \label{similar_query}
\end{figure}

In the beginning,  $\alpha$ is set to  $(1,1,1)$, which means the possibility of choosing any option is the same.  Along with the annotation process, we update the posterior distribution by the answered queries  $D$. In the neighbor areas of $ q_i$' each sampler, we choose the top $k$ closest super-pixel patches by their embedding distances to the sample and consider the queries sampled from these patches as $ q_i$'s similar queries (see Fig~\ref{similar_query} when $k=1$). We seek $qi$’s similar queries in  $D$, count the number of picking the similar queries' three options as $(m_1,m_2,m_3)$, and update $Dir(\alpha) $ by $p(\theta|\alpha,D)=p(\theta|\alpha_1+m_1,\alpha_2+m_2,\alpha_3+m_3)$. Then we calculate the expected value of choosing $a,b,c$ by the updated posterior distribution. If any option's expected picking possibility is higher than a threshold, we could conclude that it is a confident query for an annotator and reject this query.

The third step is to reject queries with less utility, which is assessed by the variance of $a,b,c$ three options' picking distribution. A query with high variance means it is still ambiguous to the annotator after he/she answered many similar queries and will be rejected.

From Fig~\ref{similar_query}, we can notice that annotators tend to make the same decisions when facing similar queries. For example, in the query $[a,b,c]$, if the annotator consider the sample $c$ as most dissimilar because $b$ and $a$ are all building patches, a similar choice of $c_1$ also likely to be made in query $[a_1,b_1,c_1]$. To further lighten the annotator's burden, we simulate more answers by these similar queries.

\section{Experiments}\label{Expriment} 
We conduct the experiments in three scenarios with real or virtual participants to show the value of our method.

We first evaluate our model in a synthetic image. As texture information is the main contributing factor in many scenarios, such as medical or aerial segmentation, a synthetic image with dominant texture information should be a good beginning to test our model. We simulate virtual annotators who always precisely respond based on given knowledge hierarchies to ensure no other variability is introduced, and compare the elicited knowledge structure with the given ones in an objective manner. To confirm that our method is able to create different hierarchical segmentation based on various perceptions, the simulated participants hold partial agreements about the latent perception.

Next, we test our method's potentials in real-world applications. Two experts with different perceptions about a histopathological image are simulated for psychometric testing.

Then, we assess our model's ability when dealing with real human participants with extra involved variations. The experiment is conducted on an aerial image.

\textit{We emphasize that the main contribution of our work is to elicit full-depth knowledge and concept relationships from experts rather than perfect pixel-wise prediction}. Therefore, the commonly used mIoU score in image segmentation will not be regarded as a metric in this work. Instead, we use dendrogram purity~\citep{heller2005bayesian} to measure the matching degree between our extract hierarchical segmentation and the ground-truth perception, and use each cluster's purity to quantify how accurate each segmented sub-area is. All the experiments use VGG19 as backbones. The margin value $m$ is set to 0.2. 

\subsection{Virtual Participant on Synthetic Image}\label{Virtual_synthetic}

\begin{figure}[htbp]
    \centering
    \subfigure[Nine different textures]{
        \includegraphics[width=0.23\textwidth]{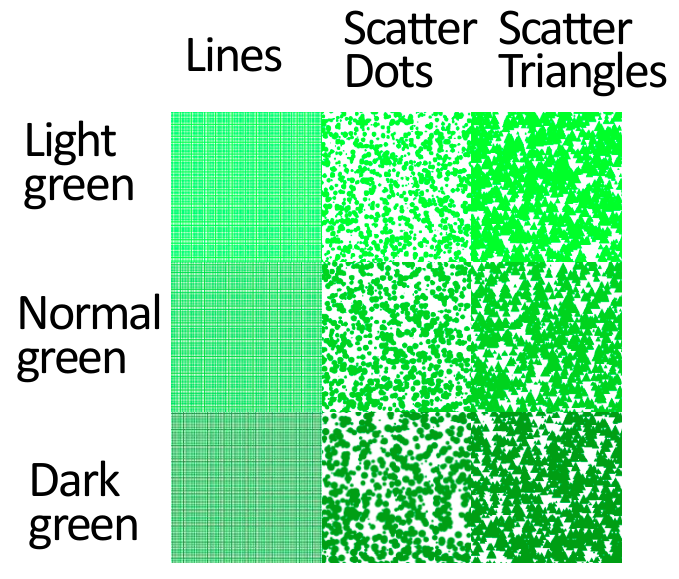}
        \label{Synthetic_image_cate}
    }
    \subfigure[Synthetic image]{
        \includegraphics[width=0.35\textwidth]{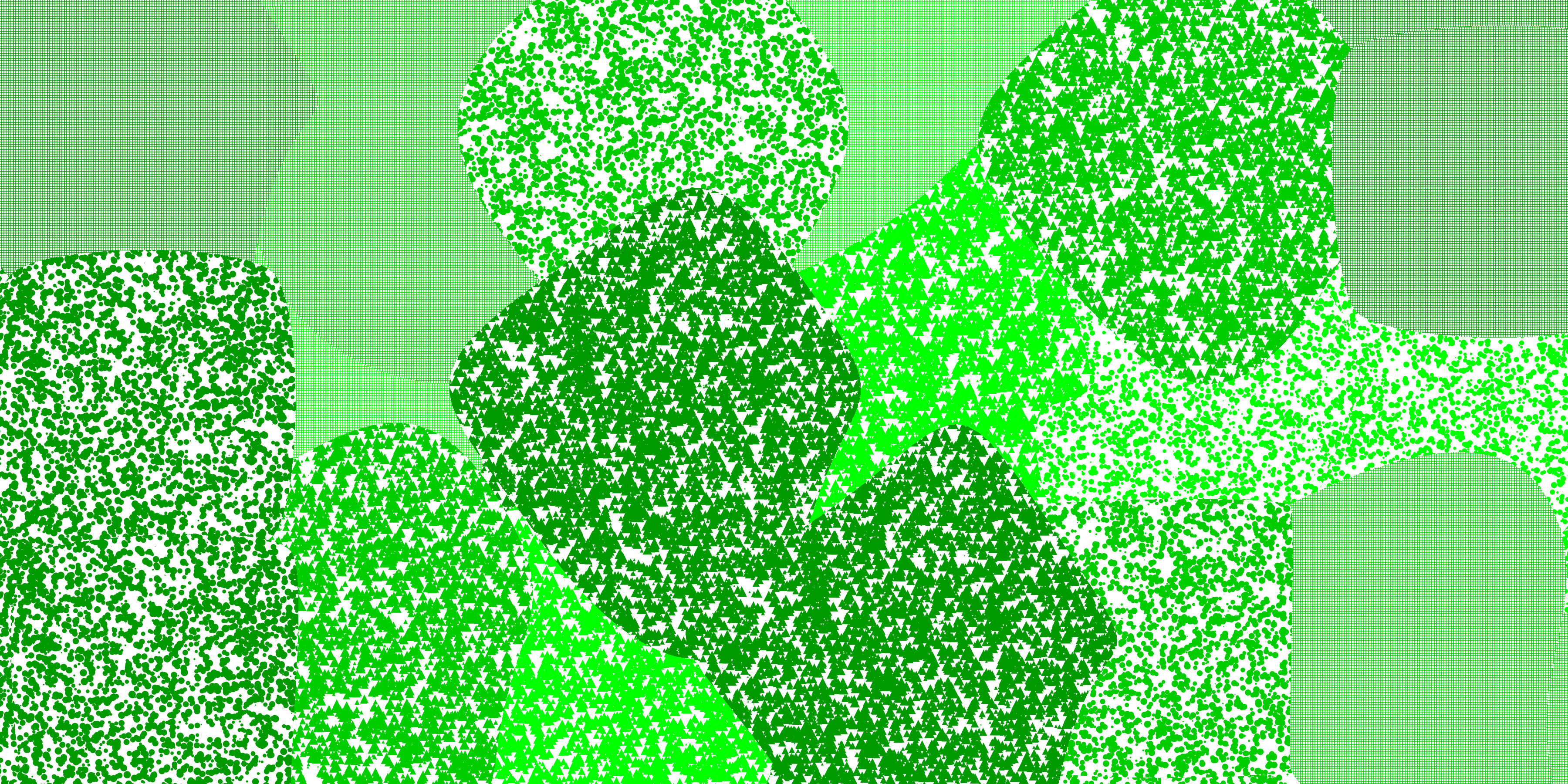}
        \label{Synthetic_image_ori}
    }
    \subfigure[Visualize the ground-truth labels ]{
	\includegraphics[width=0.35\textwidth]{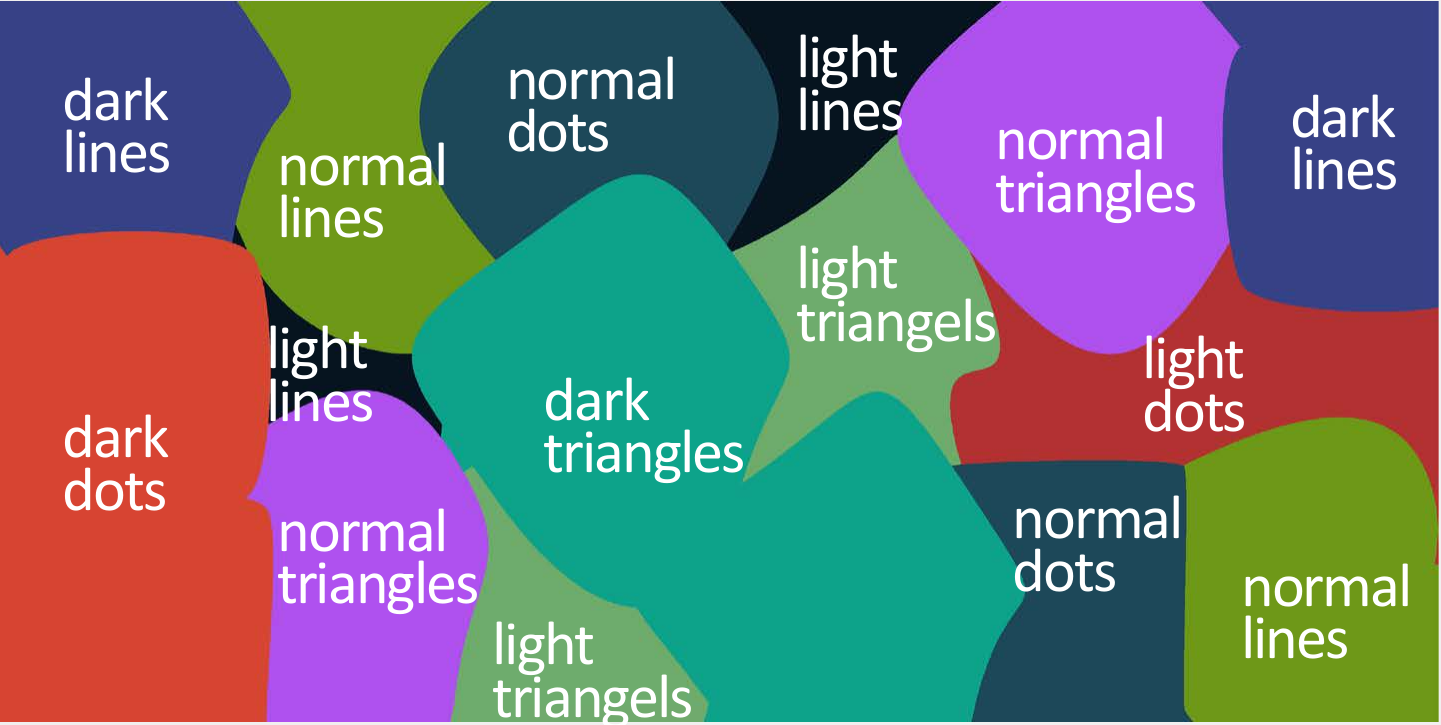}
	\label{Synthetic_image_label}
    }
    \caption{Synthetic image and the ground truth. We abbreviate dark green, normal green, light green to dark, normal, green and scatter dots, scatter triangles to dots, triangles respectively in (c) and the following figures in this paper. }
    \label{Synthetic_image}
\end{figure}

To remove other variability and test our method in a controllable way, we created a synthetic image that contains light green, normal green, dark green three colors, and lines, scatter dots, scatter triangles three textures. The nine different categories are shown in Fig~\ref{Synthetic_image}.  The whole image's  resolution is $1800\times3600\times3$ pixels.

Since human perception is not obvious before been elicited, it would be ambiguous to assess how well the extracted hierarchical segmentation matches the latent ground truth. Therefore, we create two virtual participants with given hierarchies to simulate the latent knowledge and  generate responses accordingly.

\begin{figure}[htbp]
    \centering
    \subfigure[Knowledge hierarchy and the extracted results] {
        \includegraphics[width=0.7\textwidth]{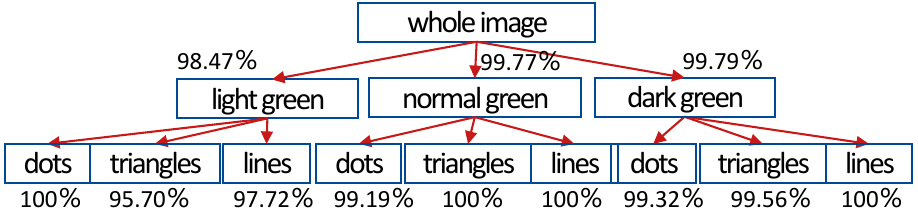}
    }
      \subfigure[Palette]{
        \includegraphics[width=0.2\textwidth]{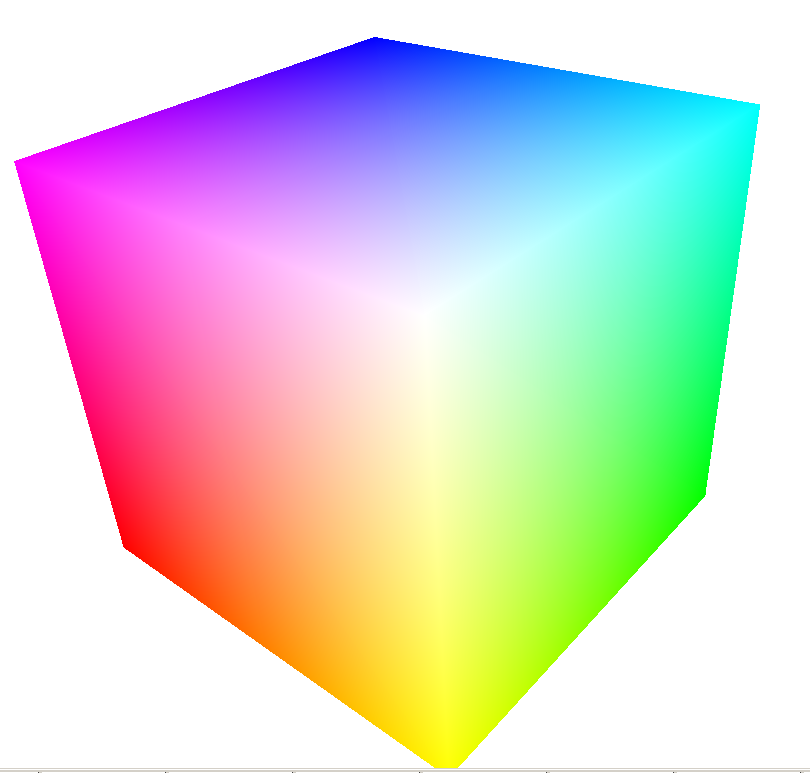}
        \label{Color_similarity}
    }
    \subfigure[Visualize segmentation at the first layer of the hierarchy]{
        \includegraphics[width=0.45\textwidth]{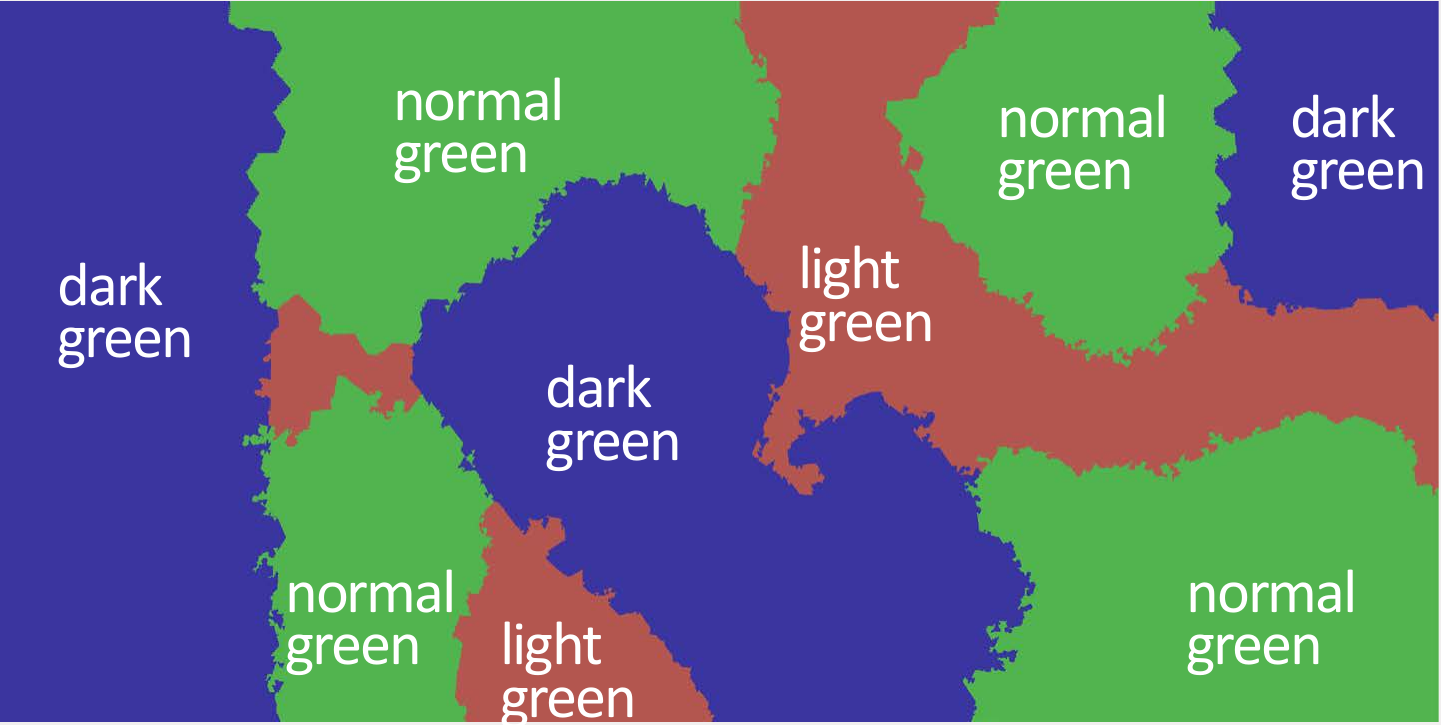}
    }
    \subfigure[Visualize segmentation at the last layer of the hierarchy
]{
	\includegraphics[width=0.45\textwidth]{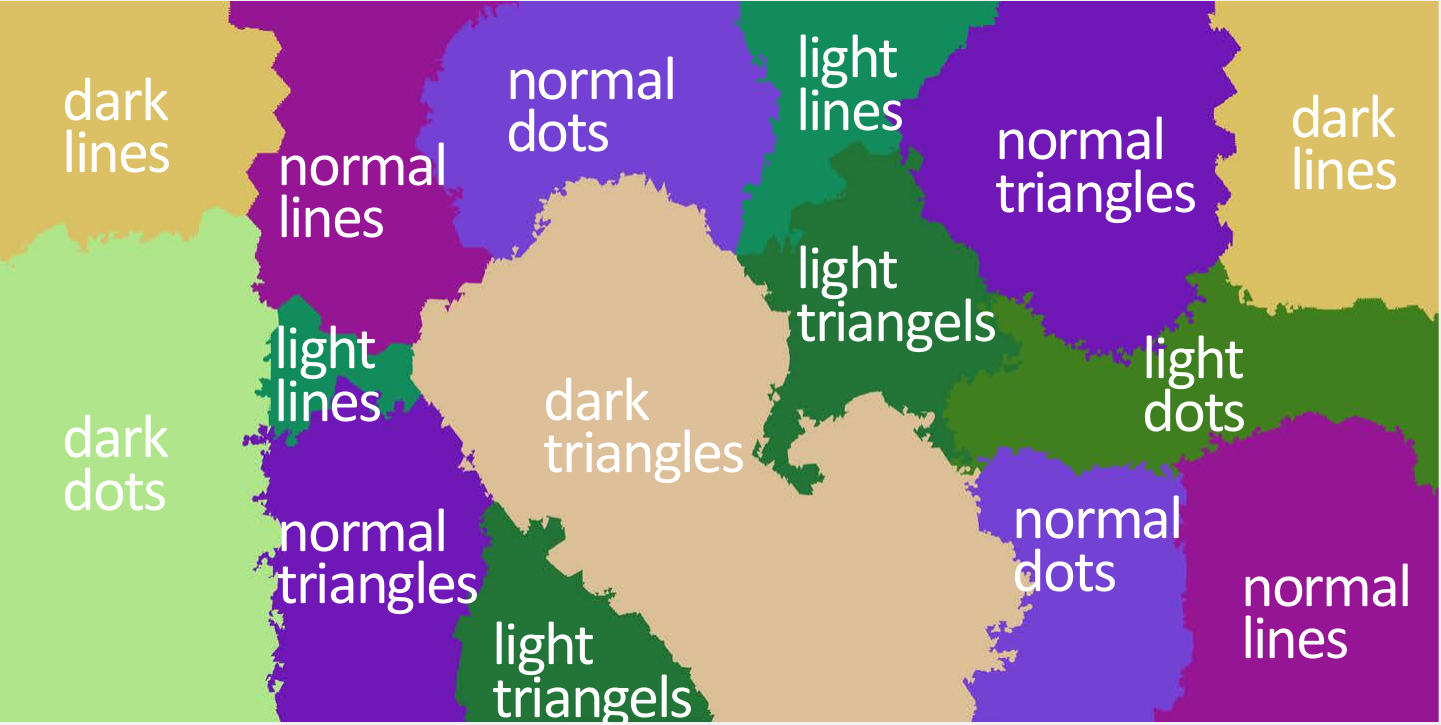}
	\label{Synthetic_results_1_2}
    }
    \caption{ (a) is a given hierarchy to simulate the latent knowledge structure from virtual participant 1. The value by each node indicates the cluster's purity extracted by our method. (b) is a palette for color similarity reference and could be used in following figures in the paper.  (c) and (d) are segmentation results in the hierarchy's first  and last layer, and  if their colors are similar based on the palette, the concepts that colors represent should also be considered close.  }

    \label{Synthetic_results_1}
\end{figure}

\begin{figure}[htbp]
\centering
    \subfigure[Knowledge hierarchy and the extracted results]{
        \includegraphics[width=0.75\textwidth]{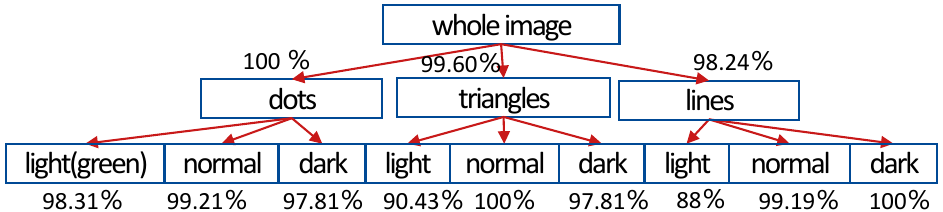}
    }
    
    \subfigure[Visualize segmentation at the first layer of the hierarchy]{
        \includegraphics[width=0.45\textwidth]{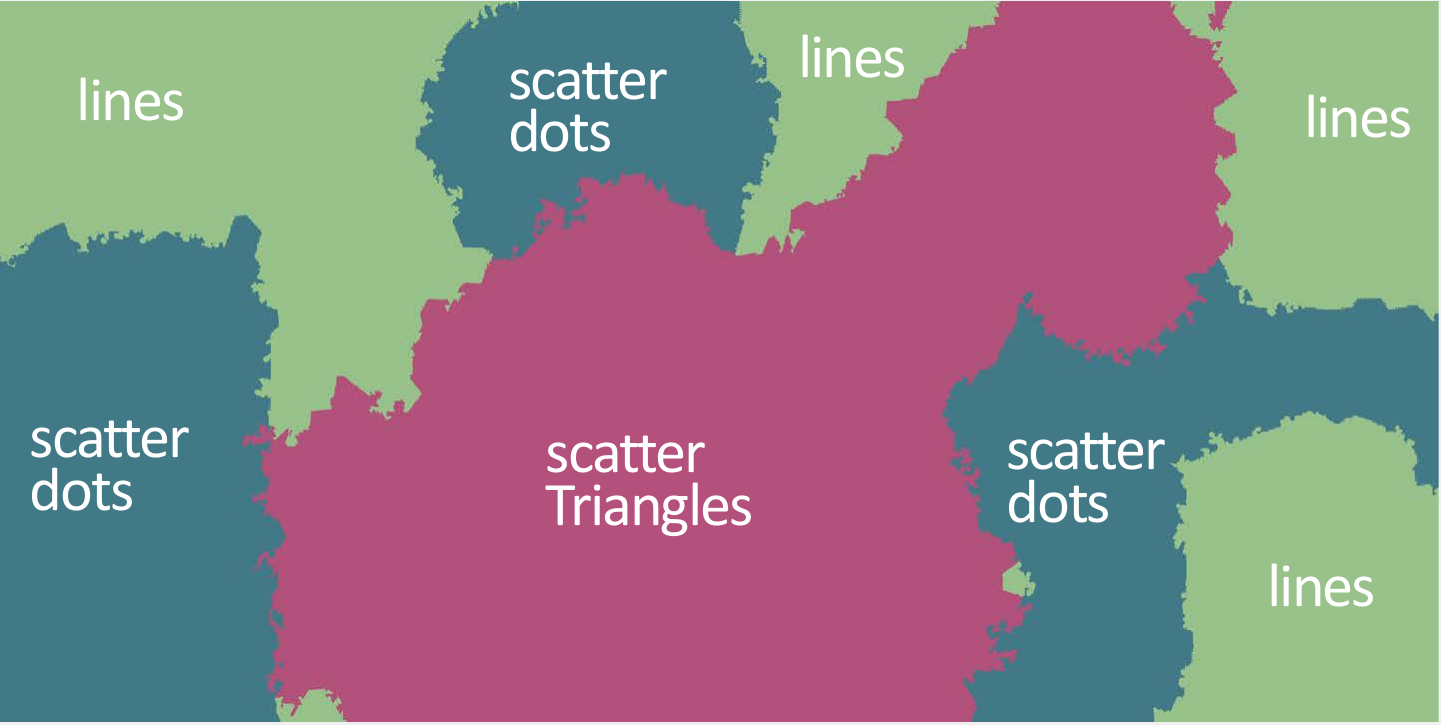}
    }
    \subfigure[Visualize segmentation at the last layer of the hierarchy
]{
	\includegraphics[width=0.45\textwidth]{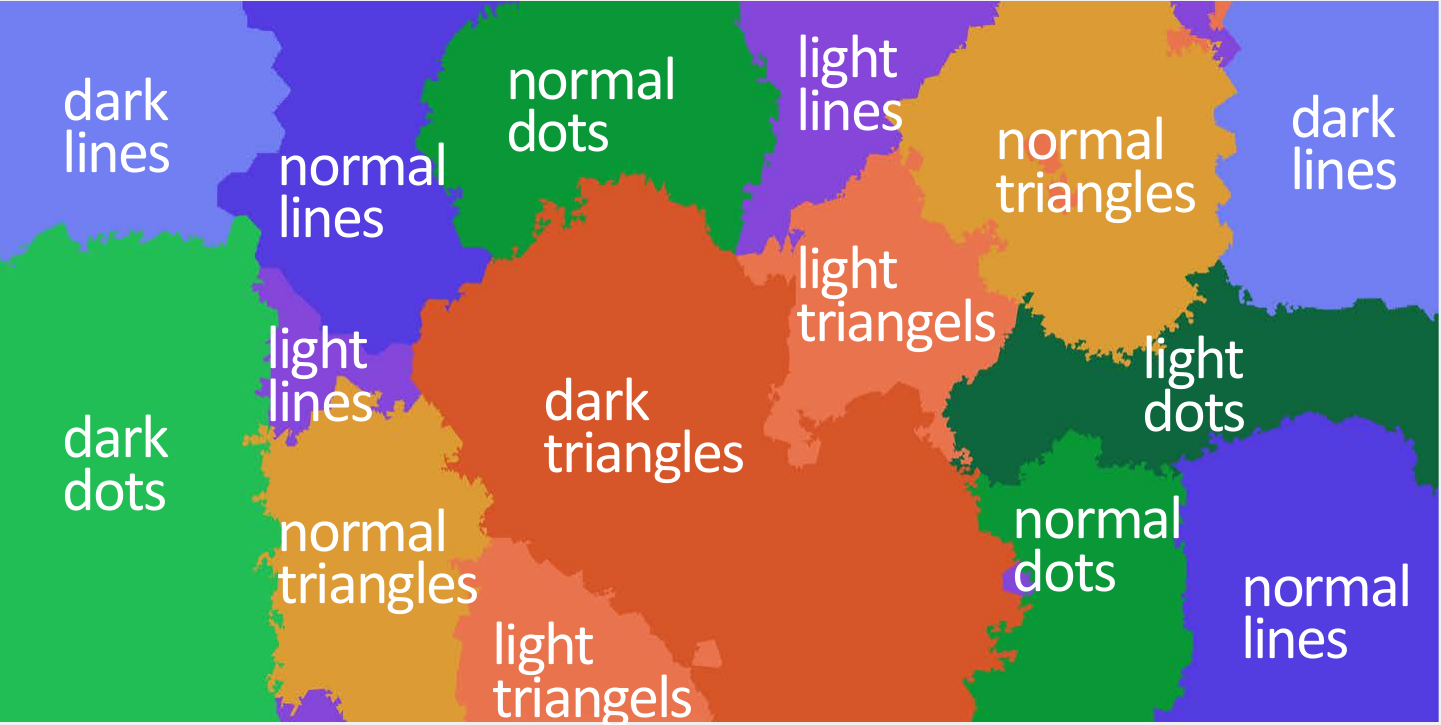}
    }
\caption{Knowledge structure from virtual participant 2 and hierarchical segmentation results on synthetic image. }
    \label{Synthetic_results_2}
\end{figure}

To be specific, we create $1112$ super-pixel patches by SLIC for psychometric tests and train the model with 10 iterations. Each iteration contains 800 simulated responses and another two times enhanced triplets.  We illustrate the first participant's segmentation result in Fig~\ref{Synthetic_results_1}.  The result shows that our model is capable of eliciting a clear knowledge structure that well matches the latent perception (simulated by given hierarchy), with all sub-cluster purity above $95\%$.  In the first-layer segmentation, results are based on the areas' color (dark green, normal green, light green). Then the segmentation splits into detailed nine categories in the last layer. Note that the color similarities match the concept relationships well in the segmentation visualization. For instance, all the normal green areas in the original image are marked as purple in Fig~\ref{Synthetic_results_1_2} as these areas sharing the same normal green color, but the purple has different shades since these areas have different textures (scatter triangles, scatter dots, lines). We can observe similar results on the second participant  (see Fig~\ref{Synthetic_results_2}), but with a different knowledge structure.  The image is segmented based on textures  on a higher layer then divided by colors.  That confirms our model's ability to create various hierarchical segmentation results based on different latent perceptions.


\subsection{ Virtual Participants on Histopathological Image}

Then we evaluate our model on the real-world histopathological image from BACH Challenge 2018~\citep{aresta2019bach}. We crop the image to $4211\times10020\times3$ pixels, create $2574$ super-pixels, and train the model with 10 iterations. 1000  responses are simulated in each iteration except 1500 responses in the first one. Answered triplets are doubled by the proposed enhancement algorithm before used for training. Two participants with different latent hierarchical knowledge are simulated.

\begin{figure}[htbp]
    \centering
    \subfigure[Knowledge hierarchy and the extracted results]{
        \includegraphics[width=0.7\textwidth]{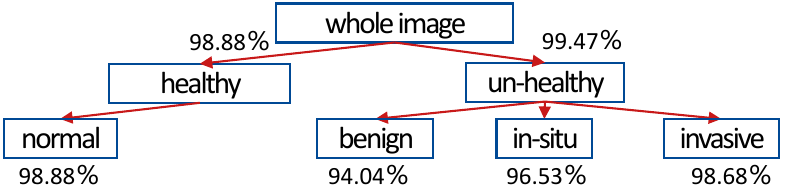}
    }
    
    \subfigure[Visualize segmentation at the first layer of the hierarchy]{
        \includegraphics[width=0.45\textwidth]{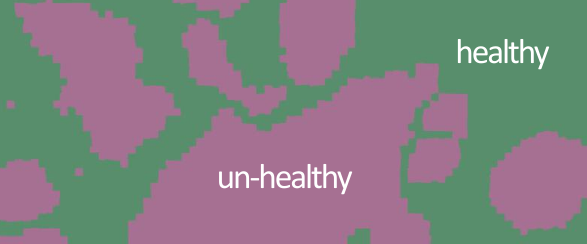}
    }
    \subfigure[Visualize segmentation at the last layer of the hierarchy
]{
	\includegraphics[width=0.45\textwidth]{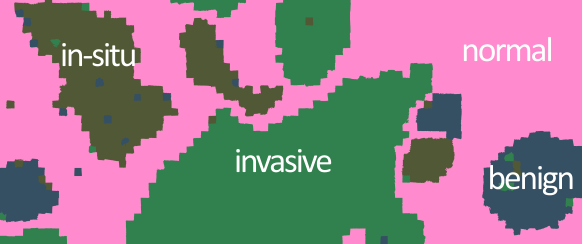}
	\label{medical_layer2_1}
    }
\caption{ Segmentation results of virtual participant 1 on histopathological image.}
    \label{medical_results_1}
\end{figure}

\begin{figure}[htbp]
    \centering
    \subfigure[Knowledge hierarchy and the extracted results]{
        \includegraphics[width=0.7\textwidth]{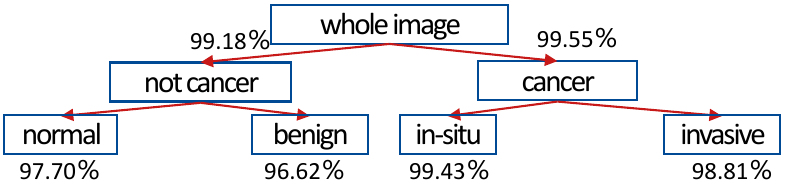}
    }
    
    \subfigure[Visualize segmentation at the first layer of the hierarchy]{
        \includegraphics[width=0.45\textwidth]{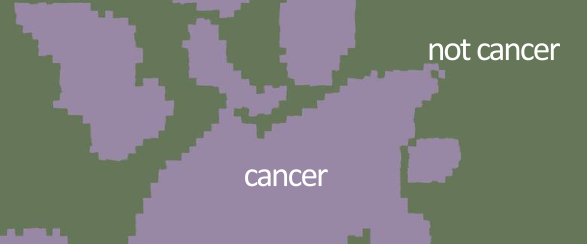}
    }
    \subfigure[Visualize segmentation at the last layer of the hierarchy
]{
	\includegraphics[width=0.455\textwidth]{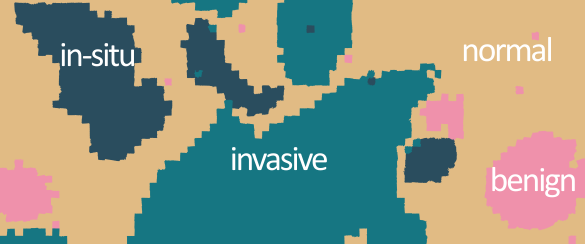}
    }
\caption{ Segmentation results of virtual participant 2 on histopathological image.}
    \label{medical_results_2}
\end{figure}

We present the elicited hierarchy and segmentation results of participant 1 in Fig~\ref{medical_results_1}. It can be seen that a clear hierarchical structure is extracted with each note's purity higher than $94\%$. The first layer of segmentation is built according to if a cell is healthy. Then it goes down to four specific categories. Besides, the colors of benign, in-situ, invasive are similar to each other but have a big difference from the normal area in visualization Fig~\ref{medical_layer2_1}, which matches the conceptual similarity in the elicited knowledge hierarchy.  The results of participant 2 are shown in  Fig~\ref{medical_results_2}. As the different knowledge tree from participant 1, the higher-level segmentation is formed based on if the area is cancer.

\begin{figure}[htbp]
    \centering
    \subfigure[Histopathological image]{
        \includegraphics[width=0.45\textwidth]{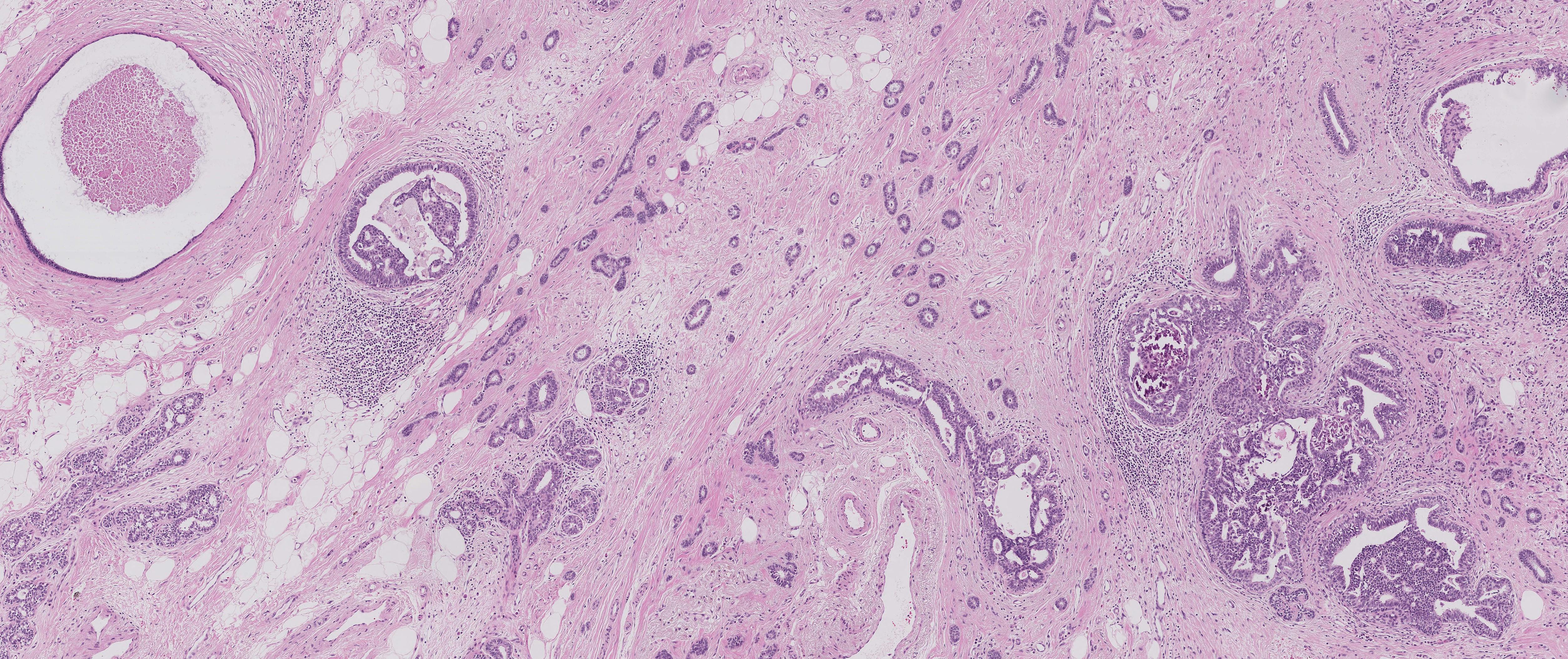}
    }
    \subfigure[Visualize the ground truth  label ]{
	\includegraphics[width=0.45\textwidth]{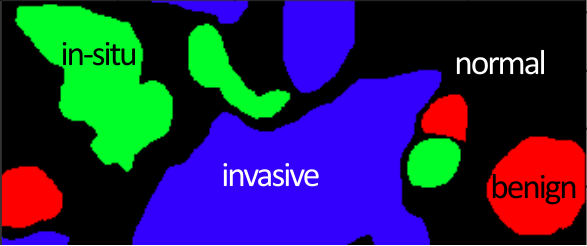}
    }
    \caption{Histopathological image and the ground truth label.}
    \label{medical_image}
\end{figure}


\subsection{ Real Participant on Aerial Image}

At last, we involved a real human in our psychometric tests. The goal is to test our model's ability to deal with extra variability introduced by a human operator.  821 suer-pixel patches are created from a $2128\times1619\times3$ aerial image\footnote{From repository https://github.com/dronedeploy/dd-ml-segmentation-benchmark} for psychometric tests.  The original image's five labels: building, clutter, vegetation, water and ground are removed and will not be presented to the participant. We collect 600 responses from the first iteration and 400 responses from the following 4 iterations.

\begin{figure}[htbp]
    \centering
    
    \subfigure[Extracted perception hierarchy and the purity]{
        \includegraphics[width=0.7\textwidth]{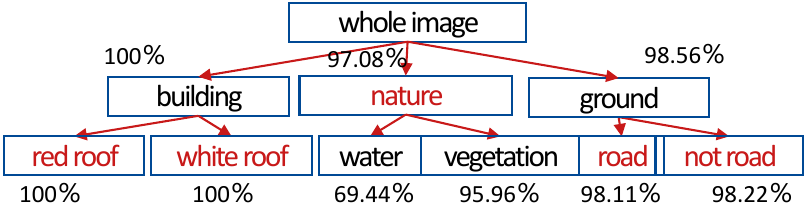}
    }
    
    \subfigure[Original image]{
        \includegraphics[width=0.31\textwidth]{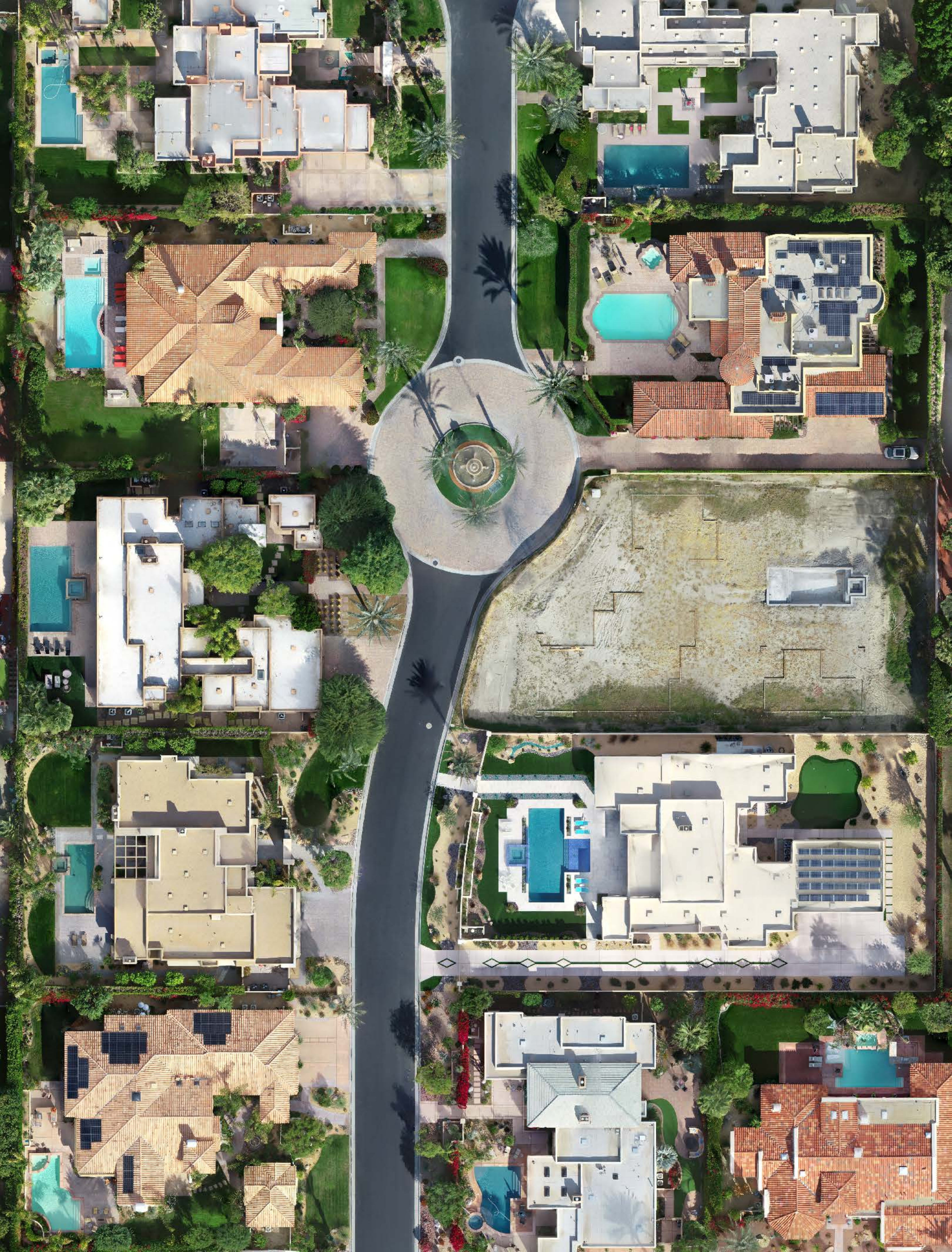}
    }
    \subfigure[Visualize segmentation at the first layer]{
        \includegraphics[width=0.31\textwidth]{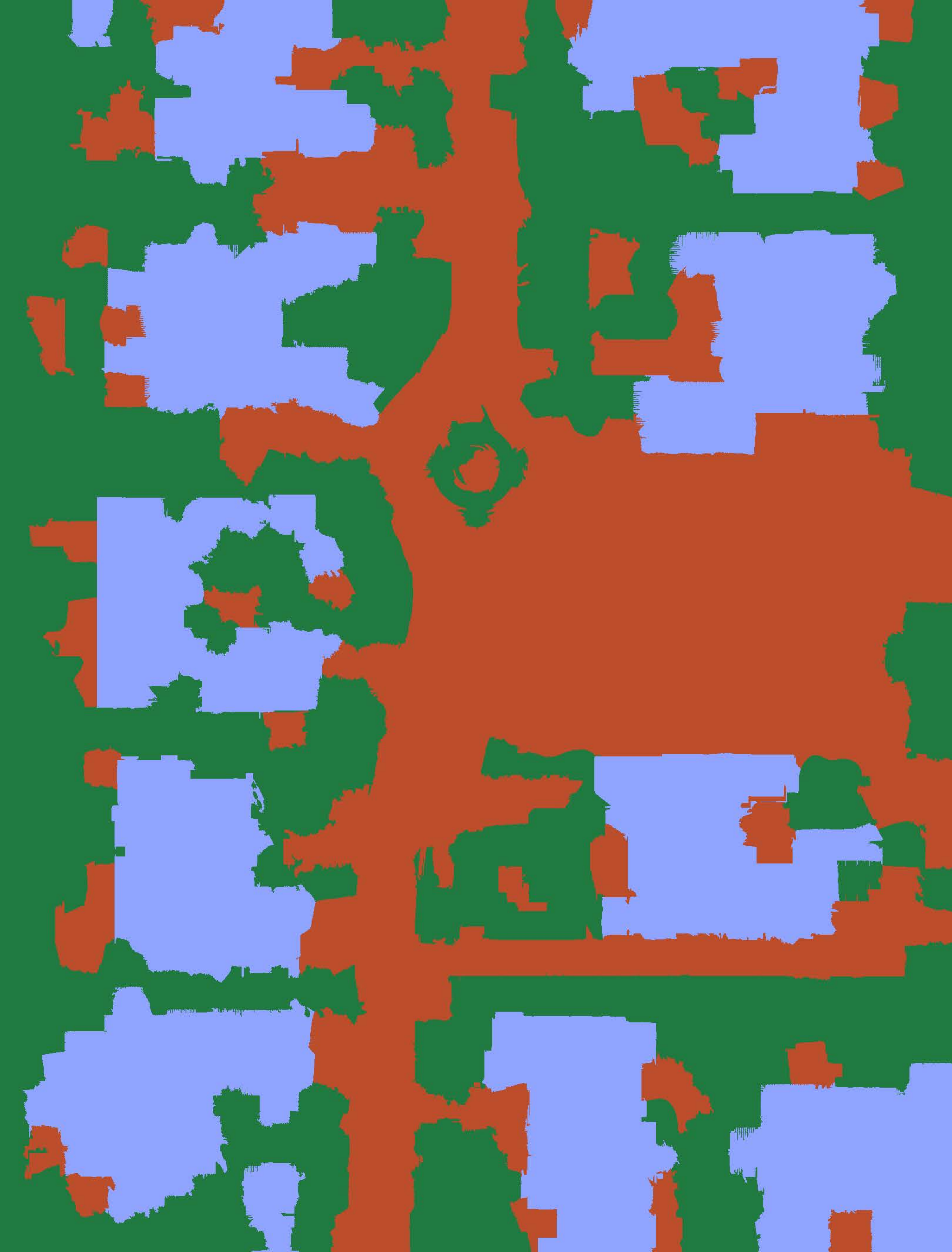}
    }
        \subfigure[Segmentation at the first layer with a mask]{
        \includegraphics[width=0.31\textwidth]{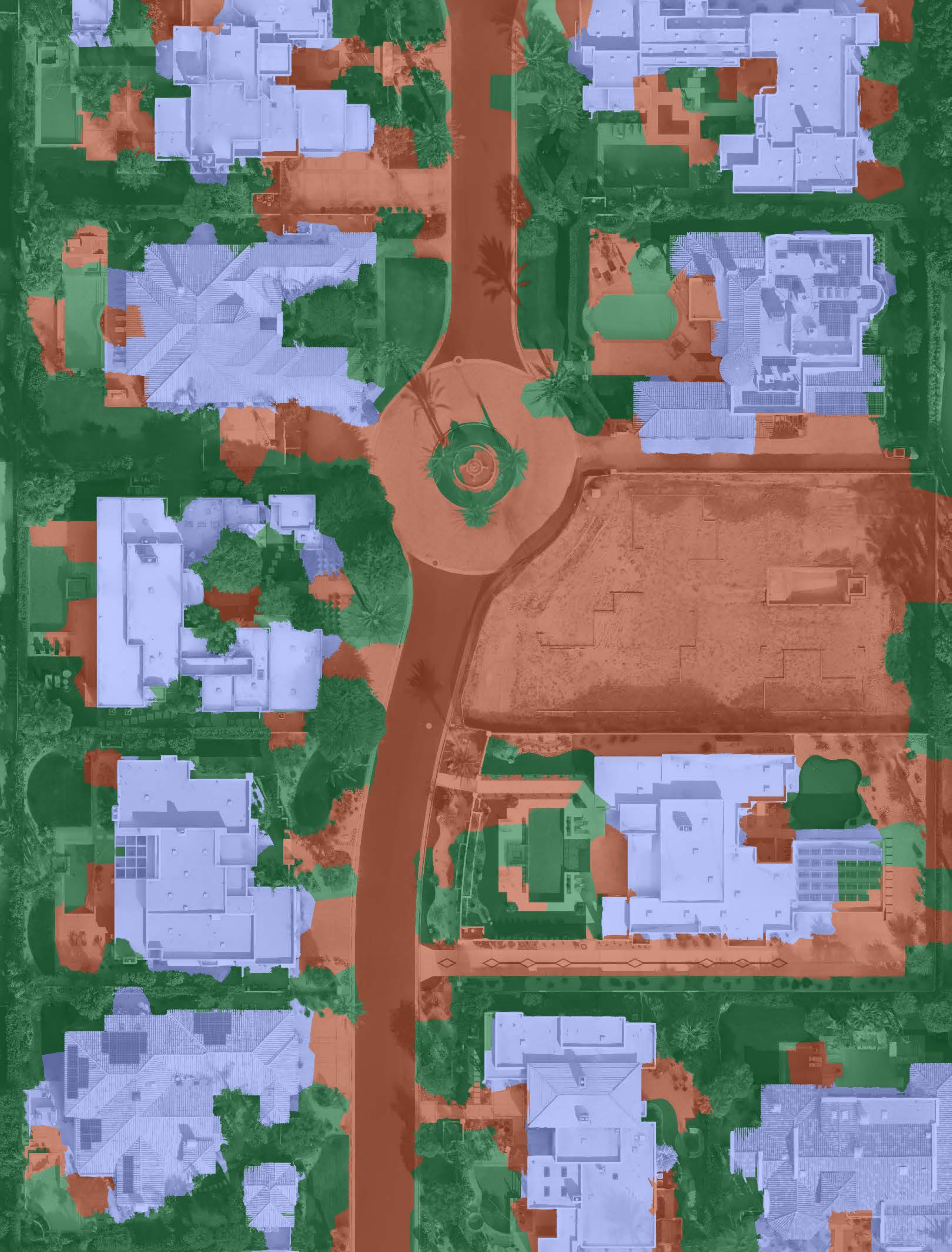}
    }

    \subfigure[Visualize segmentation at the second layer]{
        \includegraphics[width=0.31\textwidth]{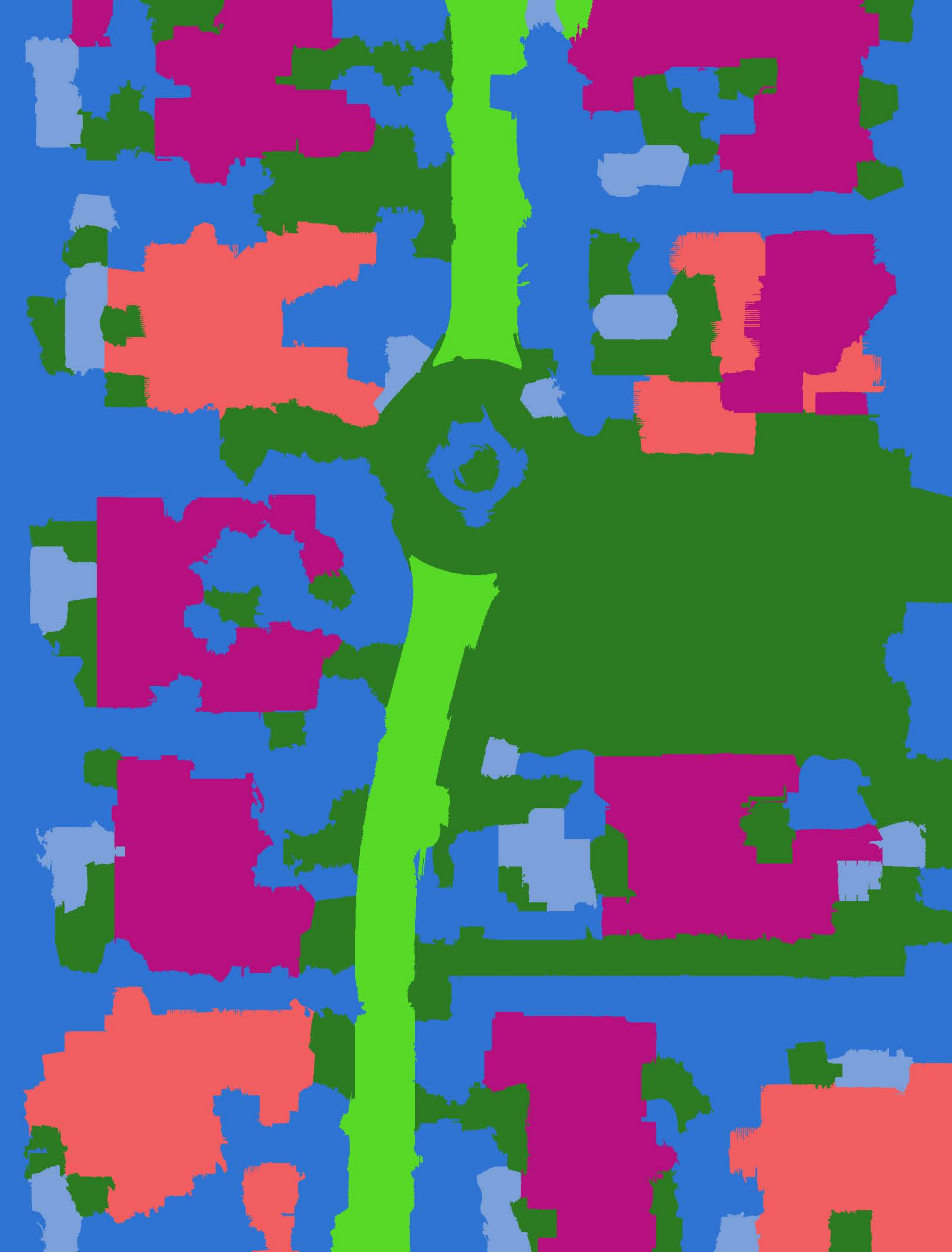}
    }
        \hspace{.3in}
        \subfigure[Segmentation at the second layer with a mask ]{
        \includegraphics[width=0.31\textwidth]{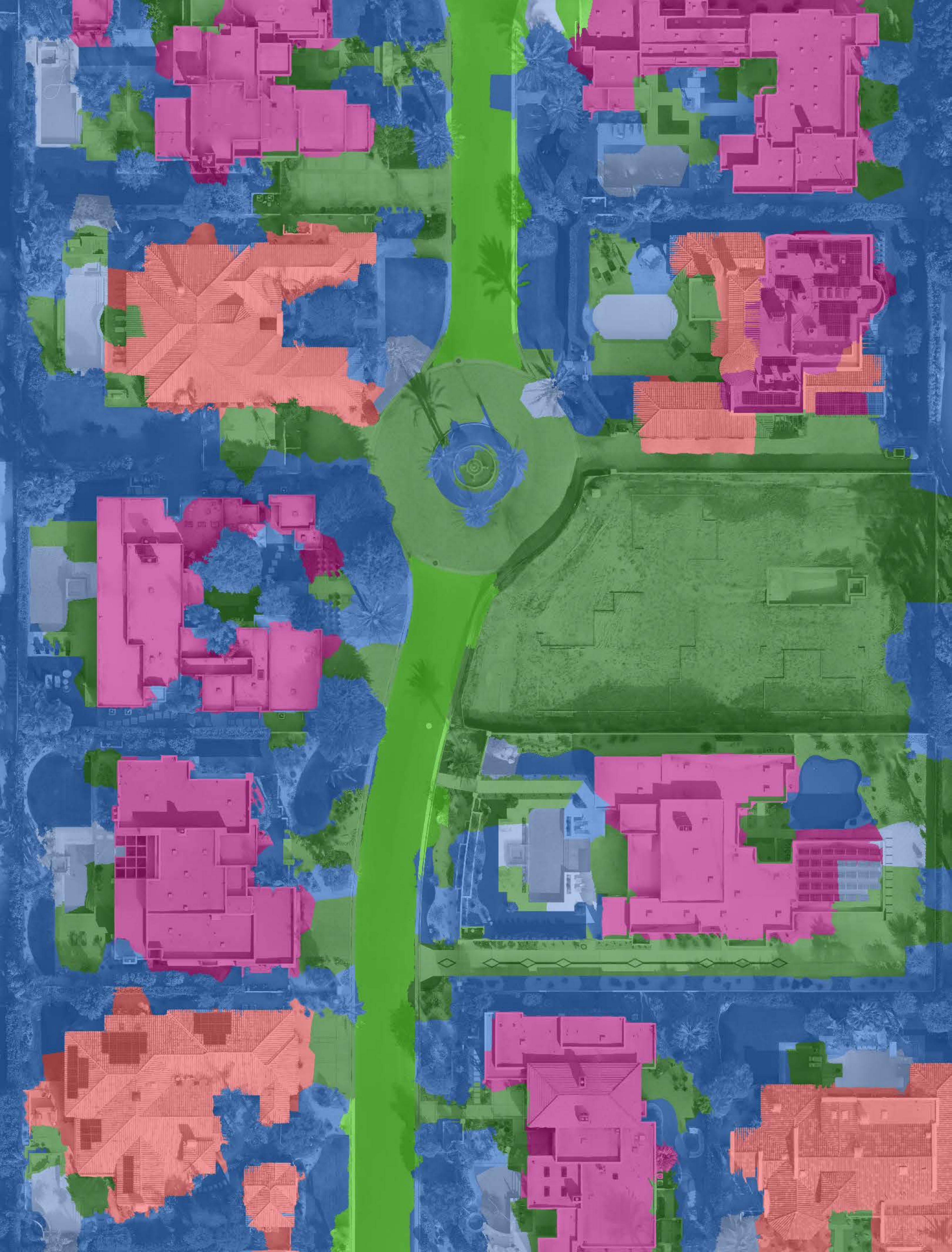}
        \label{satelite_results_f}
    }

\caption{Segmentation results of the real participant on the aerial image. (a) is the elicited knowledge structure by our method. Some new concepts are discovered beyond the original labels and are marked as red color. In (f), we could visually inspect concept relationships by colors. Close concepts are masked by similar colors and vice versa. For instance, red roof and white roof buildings are all masked red but with different color shades, because they are different categories but all belong to a bigger concept, building. }
    \label{satelite_results}
\end{figure}

We present the results in Fig~\ref{satelite_results}. A knowledge hierarchy is extracted based on the psychometric responses. We can notice that several concepts are discovered beyond the original labels, such as buildings divided into two sub-segmentation based on the colors of the roofs, the grounds split by whether vehicles could be driven on, water and vegetation merged together to natural habitats.  We manually labeled each patch with these concepts and calculated the purity of each cluster. The purities are all above  $95\%$ except the water, because the area of water is too small to generate enough super-pixels for psychometric tests. Note that in the second layer segmentation, we could tell the concept relationships directly by the color similarities. For instance, in Fig~\ref{satelite_results_f}, the color of water and vegetation are all blue because they are all natural habitats, while vegetation is dark blue and water is light blue since they are different categories.

\subsection{Comparison and Ablation Study }
Two benchmarks are designed for comparison.  First,  we label each super-pixel by a set of fixed labels (the bottom layer's labels in the hierarchy) for training and cluster on the model's activation layer to build hierarchical segmentation. Another is the hierarchical segmentation proposed by~\citet{arbelaez2010contour}. Dendrogram purity is applied to measure the quality of hierarchical segmentation against the ground truth knowledge. The results are reported in Table~\ref{Comparison_purity}. It could be noticed that on both datasets, our method could get the best performance, indicate the capability of hierarchically segmenting based on the annotator's high fidelity knowledge. It is no surprise that baseline 1 (hierarchical segmentation~\citep{arbelaez2010contour}) performs worst because no semantic perception is captured from the annotator. The results of baseline 2 (fixed label annotation) are much better but still worse than our framework since lacking the ability to extract full-depth knowledge. The gap between baseline 2 and the proposed method in histopathological image is smaller than that in synthetic image, as the hierarchical perception structure is more complex in the synthetic image, allowing our method to capture greater semantic knowledge.

\begin{table}[htbp]
\newcommand{\tabincell}[2]{\begin{tabular}{@{}#1@{}}#2\end{tabular}}  
\centering
\caption{Comparison with baselines by dendrogram purity ($\%$).}
\begin{tabular}{|c|c|c|c|c|}  

\hline
Participant & Dataset & \tabincell{c}{Baseline 1} & \tabincell{c}{Baseline 2} &\tabincell{c}{Proposed method}   \\
\hline
1 & Synthetic image & 48.86 & 71.16&  \textcolor{red}{99.83}    \\ 
2 & Synthetic image & 44.71 & 73.19&  \textcolor{red}{93.47}    \\ 
1 & Histopathological image & 47.70 & 89.14&  \textcolor{red}{94.26}    \\ 
2 & Histopathological image & 48.63 & 88.81&  \textcolor{red}{97.97}    \\ 
\hline
\end{tabular}
\label{Comparison_purity}
\end{table}

\begin{figure}[htbp]
    \centering
	\includegraphics[width=1\textwidth]{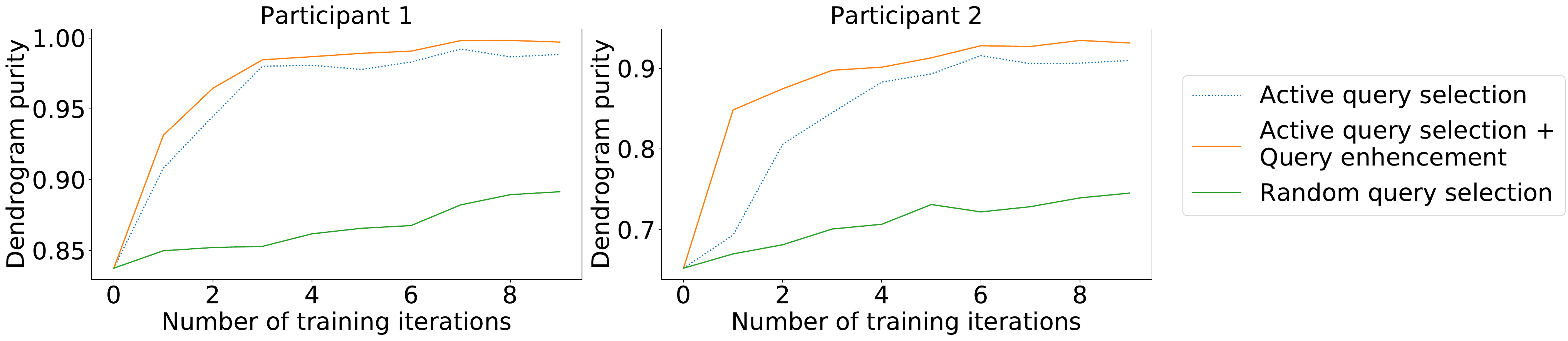}
    \caption{Ablation study on synthetic image.}
    \label{Traing_curve}
\end{figure}

To study the value of query selection and query enhancement strategies, we perform an ablation study on the synthetic image. We use the same experiment configuration as section~\ref{Virtual_synthetic}. The training curve is shown in Fig~\ref{Traing_curve}. We can see that with the same training data, our results get a noticeable boost from random selection to active queries selection and a slight improvement with the triplet enhancement strategy, indicating the effectiveness of our proposed query scheme.

\section{Conclusion}
In this work, we presented a method that captures the fine-grained semantic relationships between concepts in image segmentation using psychometric testing and deep metric learning. We empirically evaluated the ability of our method to effectively capture and represent the high fidelity semantics in synthetic, aerial and histology images. We present the results with a color overlay visualization where the captured concepts' distances are reflected by the color distances in a given palette, which allows for an effective visual inspection. The results indicate the potential of this method to improve the elicitation of knowledge for image segmentation in a broad range of image analysis applications in domains such as biology, medicine, as industrial applications.

\bibliography{ref}

\begin{thebibliography}{28}
\providecommand{\natexlab}[1]{#1}
\providecommand{\url}[1]{\texttt{#1}}
\expandafter\ifx\csname urlstyle\endcsname\relax
  \providecommand{\doi}[1]{doi: #1}\else
  \providecommand{\doi}{doi: \begingroup \urlstyle{rm}\Url}\fi

\bibitem[Achanta et~al.(2012)Achanta, Shaji, Smith, Lucchi, Fua, and
  S{\"u}sstrunk]{achanta2012slic}
Radhakrishna Achanta, Appu Shaji, Kevin Smith, Aurelien Lucchi, Pascal Fua, and
  Sabine S{\"u}sstrunk.
\newblock Slic superpixels compared to state-of-the-art superpixel methods.
\newblock \emph{IEEE transactions on pattern analysis and machine
  intelligence}, 34\penalty0 (11):\penalty0 2274--2282, 2012.

\bibitem[Arbelaez et~al.(2010)Arbelaez, Maire, Fowlkes, and
  Malik]{arbelaez2010contour}
Pablo Arbelaez, Michael Maire, Charless Fowlkes, and Jitendra Malik.
\newblock Contour detection and hierarchical image segmentation.
\newblock \emph{IEEE transactions on pattern analysis and machine
  intelligence}, 33\penalty0 (5):\penalty0 898--916, 2010.

\bibitem[Aresta et~al.(2019)Aresta, Ara{\'u}jo, Kwok, Chennamsetty, Safwan,
  Alex, Marami, Prastawa, Chan, Donovan, et~al.]{aresta2019bach}
Guilherme Aresta, Teresa Ara{\'u}jo, Scotty Kwok, Sai~Saketh Chennamsetty,
  Mohammed Safwan, Varghese Alex, Bahram Marami, Marcel Prastawa, Monica Chan,
  Michael Donovan, et~al.
\newblock Bach: Grand challenge on breast cancer histology images.
\newblock \emph{Medical image analysis}, 56:\penalty0 122--139, 2019.

\bibitem[Chen et~al.(2018)Chen, Zhu, Papandreou, Schroff, and
  Adam]{chen2018encoder}
Liang-Chieh Chen, Yukun Zhu, George Papandreou, Florian Schroff, and Hartwig
  Adam.
\newblock Encoder-decoder with atrous separable convolution for semantic image
  segmentation.
\newblock In \emph{Proceedings of the European conference on computer vision
  (ECCV)}, pages 801--818, 2018.

\bibitem[Chen et~al.(2017)Chen, Chen, Zhang, and Huang]{chen2017beyond}
Weihua Chen, Xiaotang Chen, Jianguo Zhang, and Kaiqi Huang.
\newblock Beyond triplet loss: a deep quadruplet network for person
  re-identification.
\newblock In \emph{Proceedings of the IEEE conference on computer vision and
  pattern recognition}, pages 403--412, 2017.

\bibitem[DeCarlo(2012)]{decarlo2012signal}
Lawrence~T DeCarlo.
\newblock On a signal detection approach to m-alternative forced choice with
  bias, with maximum likelihood and bayesian approaches to estimation.
\newblock \emph{Journal of Mathematical Psychology}, 56\penalty0 (3):\penalty0
  196--207, 2012.

\bibitem[Fechner(1860)]{fechner1860elemente}
Gustav~Theodor Fechner.
\newblock \emph{Elemente der psychophysik}, volume~2.
\newblock Breitkopf u. H{\"a}rtel, 1860.

\bibitem[Feng et~al.(2014)Feng, Marcellin, and Bilgin]{feng2014methodology}
Hsin-Chang Feng, Michael~W Marcellin, and Ali Bilgin.
\newblock A methodology for visually lossless jpeg2000 compression of
  monochrome stereo images.
\newblock \emph{IEEE Transactions on Image Processing}, 24\penalty0
  (2):\penalty0 560--572, 2014.

\bibitem[Fu et~al.(2019)Fu, Liu, Tian, Li, Bao, Fang, and Lu]{fu2019dual}
Jun Fu, Jing Liu, Haijie Tian, Yong Li, Yongjun Bao, Zhiwei Fang, and Hanqing
  Lu.
\newblock Dual attention network for scene segmentation.
\newblock In \emph{Proceedings of the IEEE/CVF Conference on Computer Vision
  and Pattern Recognition}, pages 3146--3154, 2019.

\bibitem[Gescheider(2013)]{gescheider2013psychophysics}
George~A Gescheider.
\newblock \emph{Psychophysics: the fundamentals}.
\newblock Psychology Press, 2013.

\bibitem[Ghodsi(2006)]{ghodsi2006dimensionality}
Ali Ghodsi.
\newblock Dimensionality reduction a short tutorial.
\newblock \emph{Department of Statistics and Actuarial Science, Univ. of
  Waterloo, Ontario, Canada}, 37\penalty0 (38):\penalty0 2006, 2006.

\bibitem[Hadsell et~al.(2006)Hadsell, Chopra, and
  LeCun]{hadsell2006dimensionality}
Raia Hadsell, Sumit Chopra, and Yann LeCun.
\newblock Dimensionality reduction by learning an invariant mapping.
\newblock In \emph{2006 IEEE Computer Society Conference on Computer Vision and
  Pattern Recognition (CVPR'06)}, volume~2, pages 1735--1742. IEEE, 2006.

\bibitem[Heller and Ghahramani(2005)]{heller2005bayesian}
Katherine~A Heller and Zoubin Ghahramani.
\newblock Bayesian hierarchical clustering.
\newblock In \emph{Proceedings of the 22nd international conference on Machine
  learning}, pages 297--304, 2005.

\bibitem[Jogan and Stocker(2014)]{jogan2014new}
Matja{\v{z}} Jogan and Alan~A Stocker.
\newblock A new two-alternative forced choice method for the unbiased
  characterization of perceptual bias and discriminability.
\newblock \emph{Journal of Vision}, 14\penalty0 (3):\penalty0 20--20, 2014.

\bibitem[Kaya and Bilge(2019)]{kaya2019deep}
Mahmut Kaya and Hasan~{\c{S}}akir Bilge.
\newblock Deep metric learning: A survey.
\newblock \emph{Symmetry}, 11\penalty0 (9):\penalty0 1066, 2019.

\bibitem[Khoreva et~al.(2017)Khoreva, Benenson, Hosang, Hein, and
  Schiele]{khoreva2017simple}
Anna Khoreva, Rodrigo Benenson, Jan Hosang, Matthias Hein, and Bernt Schiele.
\newblock Simple does it: Weakly supervised instance and semantic segmentation.
\newblock In \emph{Proceedings of the IEEE conference on computer vision and
  pattern recognition}, pages 876--885, 2017.

\bibitem[Kodinariya and Makwana(2013)]{kodinariya2013review}
Trupti~M Kodinariya and Prashant~R Makwana.
\newblock Review on determining number of cluster in k-means clustering.
\newblock \emph{International Journal}, 1\penalty0 (6):\penalty0 90--95, 2013.

\bibitem[Liu et~al.(2015)Liu, Rabinovich, and Berg]{liu2015parsenet}
Wei Liu, Andrew Rabinovich, and Alexander~C Berg.
\newblock Parsenet: Looking wider to see better.
\newblock \emph{arXiv preprint arXiv:1506.04579}, 2015.

\bibitem[Long et~al.(2015)Long, Shelhamer, and Darrell]{long2015fully}
Jonathan Long, Evan Shelhamer, and Trevor Darrell.
\newblock Fully convolutional networks for semantic segmentation.
\newblock In \emph{Proceedings of the IEEE conference on computer vision and
  pattern recognition}, pages 3431--3440, 2015.

\bibitem[Rokach and Maimon(2005)]{rokach2005clustering}
Lior Rokach and Oded Maimon.
\newblock Clustering methods.
\newblock In \emph{Data mining and knowledge discovery handbook}, pages
  321--352. Springer, 2005.

\bibitem[Schroff et~al.(2015)Schroff, Kalenichenko, and
  Philbin]{schroff2015facenet}
Florian Schroff, Dmitry Kalenichenko, and James Philbin.
\newblock Facenet: A unified embedding for face recognition and clustering.
\newblock In \emph{Proceedings of the IEEE conference on computer vision and
  pattern recognition}, pages 815--823, 2015.

\bibitem[Settles(2009)]{settles2009active}
Burr Settles.
\newblock Active learning literature survey.
\newblock 2009.

\bibitem[Sohn(2016)]{sohn2016improved}
Kihyuk Sohn.
\newblock Improved deep metric learning with multi-class n-pair loss objective.
\newblock In \emph{Proceedings of the 30th International Conference on Neural
  Information Processing Systems}, pages 1857--1865, 2016.

\bibitem[Son et~al.(2006)Son, Winslow, Yazici, and Xu]{son2006x}
IY~Son, M~Winslow, B~Yazici, and XG~Xu.
\newblock X-ray imaging optimization using virtual phantoms and computerized
  observer modelling.
\newblock \emph{Physics in Medicine \& Biology}, 51\penalty0 (17):\penalty0
  4289, 2006.

\bibitem[Suh et~al.(2019)Suh, Han, Kim, and Lee]{suh2019stochastic}
Yumin Suh, Bohyung Han, Wonsik Kim, and Kyoung~Mu Lee.
\newblock Stochastic class-based hard example mining for deep metric learning.
\newblock In \emph{Proceedings of the IEEE/CVF Conference on Computer Vision
  and Pattern Recognition}, pages 7251--7259, 2019.

\bibitem[Vernaza and Chandraker(2017)]{vernaza2017learning}
Paul Vernaza and Manmohan Chandraker.
\newblock Learning random-walk label propagation for weakly-supervised semantic
  segmentation.
\newblock In \emph{Proceedings of the IEEE conference on computer vision and
  pattern recognition}, pages 7158--7166, 2017.

\bibitem[Wang et~al.(2020)Wang, Zhang, Kan, Shan, and Chen]{wang2020self}
Yude Wang, Jie Zhang, Meina Kan, Shiguang Shan, and Xilin Chen.
\newblock Self-supervised equivariant attention mechanism for weakly supervised
  semantic segmentation.
\newblock In \emph{Proceedings of the IEEE/CVF Conference on Computer Vision
  and Pattern Recognition}, pages 12275--12284, 2020.

\bibitem[Yin et~al.(2020)Yin, Menkovski, and Pechenizkiy]{yin2020knowledge}
Lu~Yin, Vlado Menkovski, and Mykola Pechenizkiy.
\newblock Knowledge elicitation using deep metric learning and psychometric
  testing.
\newblock \emph{arXiv preprint arXiv:2004.06353}, 2020.

\end{thebibliography}

\appendix

\section{Comparison of different margin settings}\label{appendix}

\begin{figure}[htbp]
    \centering 

    \includegraphics[width=0.7\textwidth]{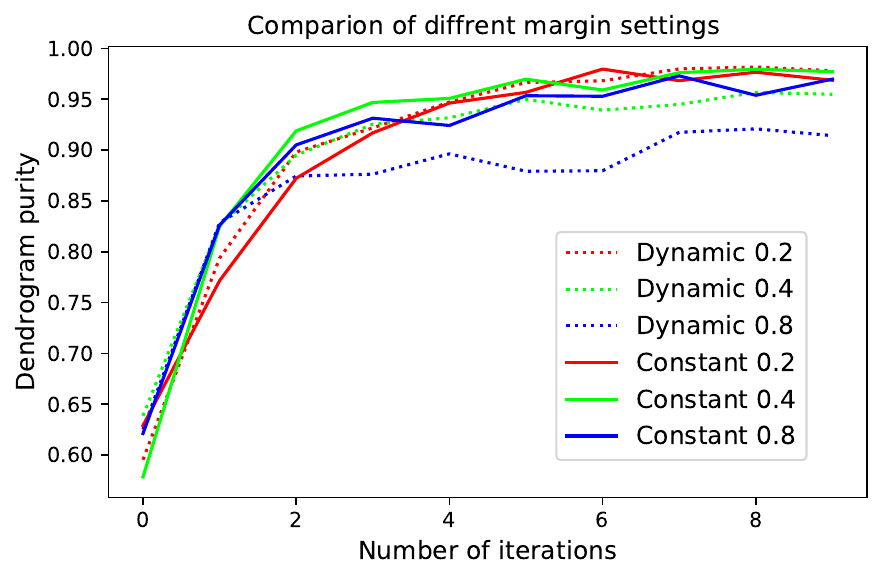} 
    
    \vspace{-0.5cm}
    \caption{Training curve with different margin settings on histopathological image
 } 
    \label{margin}
\end{figure}

To measure the impact of different margin settings, we train the model 10 iterations on a histopathological image. 1000  responses are simulated in the first iteration, and 1500 responses are simulated in each following one. Answered triplets are doubled by the proposed enhancement algorithm before feeding them into the model.  Dendrogram purity is applied to measure the performance. We applied different margin values with constant and dynamic \citep{yin2020knowledge}  settings. Results are reported in Table~\ref{Comparison_margin} and Fig~\ref{margin}. 

It could be seen that there is no much difference in performance under constant and dynamic margin settings when margin value is small, but a noticeable performance drops under dynamic 0.8 margin. That is because the extra variables introduced by dynamic margin may destabilize the model, and a large margin value could enlarge this impact.  Therefore, in this paper, we choose a constant margin with value 0.2.

\vspace{-0.8cm}

\begin{table}[htbp]
\newcommand{\tabincell}[2]{\begin{tabular}{@{}#1@{}}#2\end{tabular}}  
\centering
\caption{ Dendrogram purity of different margin settings on histopathological image ($\%$) }

\begin{tabular}{|c|c|c|c|}

\hline
\multirow{2}{*}{Iterations} & \multicolumn{3}{c|}{  \tabincell{c}{ Margin value \\ Constant/Dynamic }   }  \\
\cline{2-4}
  &   0.   2 & 0.4 & 0.8  \\
\hline
0  & 62.90/59.54 & 57.86/63.87 & 62.12/62.51  \\
\hline
1  & 77.16/79.37 & 82.48/82.61 & 82.63/82.85 \\
\hline
2  & 87.22/89.81 & 91.89/89.50 & 90.51/87.44  \\
\hline
3  & 91.69/92.18 & 94.68/92.55 & 93.13/87.63  \\
\hline
4  & 94.62/94.72 & 95.08/93.17 & 92.42/89.63  \\
\hline
5  & 95.68/96.64 & 96.97/94.98 & 95.34/87.92  \\
\hline
6  & 97.97/96.81 & 95.91/93.92 & 95.29/87.96 \\
\hline
7  & 96.82/97.98 & 97.61/94.48 & 97.30/91.74  \\
\hline
8  & 97.66/98.17 & 97.95/95.67 & 95.39/92.08 \\
\hline
9  & 96.86/97.80 & 97.71/95.49& 96.98/91.42  \\
\hline
\end{tabular}
\label{Comparison_margin}
\end{table}

\end{document}